\documentclass[10pt,twocolumn,letterpaper]{article}

\usepackage{cvpr}
\usepackage{times}
\usepackage{epsfig}
\usepackage{graphicx}
\usepackage{amsmath}
\usepackage{amssymb}
\usepackage{subfigure}
\usepackage{array}
\usepackage{multirow}
\usepackage{diagbox}
\usepackage{graphicx}
\usepackage{booktabs}
\usepackage{gensymb}
\usepackage{xspace}

\usepackage[breaklinks=true,bookmarks=false]{hyperref}
\usepackage{cleveref}

\newcommand{\OURS}{Im2Pano3D\xspace}
\newcommand*{\Cdot}{\raisebox{-0.25ex}{\scalebox{1.75}{$\cdot$}}}
\newcommand{\mypara}{\vspace*{-3mm}\paragraph}

\cvprfinalcopy 

\def\cvprPaperID{2567} 

\ifcvprfinal\fi
\begin{document}


\title{
\OURS: \\Extrapolating 360 $\degree$ Structure and Semantics Beyond the Field of View
}

\author{{Shuran Song$^{1}$  ~ Andy Zeng$^{1}$~ Angel X. Chang$^{1}$~    Manolis Savva$^{1}$ ~ Silvio Savarese$^{2}$ ~ Thomas Funkhouser$^{1}$ } \\ 
{ $^{1}$Princeton University  \quad\quad\quad $^{2}$Stanford University } \\ \href{http://im2pano3d.cs.princeton.edu}{http://im2pano3d.cs.princeton.edu}}


\maketitle


\begin{abstract}
We present \OURS, a convolutional neural network that generates a dense prediction of 3D structure and a probability distribution of semantic labels for a full $360\degree$ panoramic view of an indoor scene when given only a partial observation ($\leq 50\%$) in the form of an RGB-D image.
To make this possible, \OURS leverages strong contextual priors learned from large-scale synthetic and real-world indoor scenes. 
To ease the prediction of 3D structure, we propose to parameterize 3D surfaces with their plane equations and train the model to predict these parameters directly. 
To provide meaningful training supervision, we use multiple loss functions that consider both pixel level accuracy and global context consistency.
Experiments demonstrate that \OURS is able to predict the semantics and 3D structure of the unobserved scene with more than 56\% pixel accuracy and less than 0.52m average distance error, which is significantly better than alternative approaches. 

\end{abstract}

\section{Introduction}
People possess an incredible ability to infer contextual information from a single image \cite{intraub1989wide}. Whether it is by using prior experience or by leveraging visual cues \cite{bar2004visual,lyle2006importing}, people are adept at reasoning about what may lie beyond the field of view and make use of that information for building a coherent perception of the world \cite{hochberg1978perception}.
Similarly, in robotics and computer vision, extrapolating useful information outside a camera's field of view (FOV) plays an important role for many applications, such as goal-driven navigation\cite{zhu2017target,borenstein1989real} or next-best-view approximation \cite{jayaraman2017learning}, where a global representation of the environment can improve preemptive planning for intelligent systems. 

\begin{figure}[t]
  \vspace{-3mm}
\includegraphics[width=\linewidth]{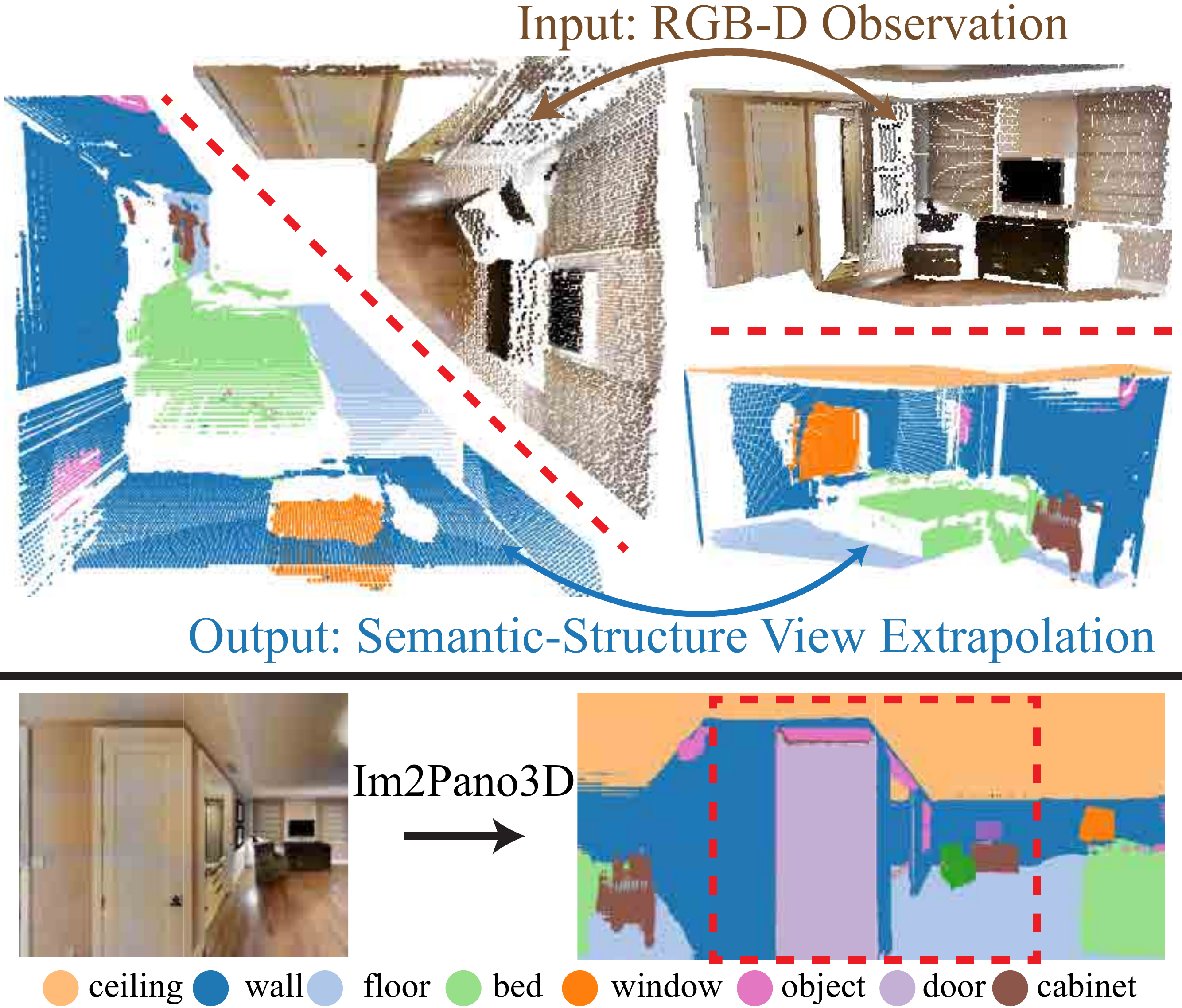}
\caption{{\bf Semantic-structure view extrapolation.} Given a partial observation of the room in the form of an RGB-D image, our \OURS  predicts both 3D structure and semantics for a full panoramic view of the same scene.}\label{fig:task}
\vspace{-3mm}
\end{figure}
However, prior work in view extrapolation typically only predicts the color pixels beyond the image boundaries  \cite{pathakCVPR16context,zhang2013framebreak,shan2014photo}. While inspiring, these methods do not predict 3D structure or semantics, and hence cannot be used directly for high-level reasoning tasks in robotics applications. 

In this paper, we explore the task of directly extrapolating 3D structure and semantics for a full panoramic view of a scene when given a view covering 50\% or less as input.   We refer to this task as {\bf semantic-structure view extrapolation}. Our method, \OURS, takes in a partial view of an indoor scene (e.g., a few RGB-D images) and uses a convolutional neural network to generate dense predictions for 3D structure and a probability distribution of semantic labels for a full $360\degree$ panoramic view of that same scene.

This is a very challenging task.  However, by learning the statistics of many typical room layouts, we can train a data-driven model to leverage contextual cues to predict what is beyond the field of view for typical indoor environments. 
For example, as shown in Fig.\ref{fig:task}, given half of a bedroom (180 \degree horizontal field of view), the system can predict the 3D structure and semantics for the other half.   This requires it not only to extend the partially observed room structures (walls, floor, ceiling, etc.), but also to predict the existence and locations of objects that are not directly observed in the input (bed, window and cabinet) using statistical properties learned from data. 

Semantic-structure view extrapolation poses three main challenges, which we address with corresponding key ideas shaping our approach to the task: 
\vspace{-1mm}
\begin{itemize}
\setlength{\topsep}{0pt}
\setlength{\parsep}{0pt}
\setlength{\parskip}{0pt}
\setlength{\itemsep}{4pt}
\item[$\Cdot$] {How to leverage strong contextual priors for indoor environments.}
\item[$\Cdot$]{How to represent the 3D structure in a way that is good not only for recognition but also for reconstruction.}
\item[$\Cdot$]{How to design meaningful loss functions when the possible solution is not unique -- a small change to object locations may still result in a valid solution.}
\end{itemize}
\vspace{-1mm}

To leverage strong \emph{contextual priors} for indoor environments, we represent 3D scenes in a single panorama image with channels encoding 3D structure and semantics. We train our model over a large-scale synthetic (SUNCG \cite{SSCNet}) and real-world indoor scenes (Matterport3D \cite{Matterport3D}) encoded in this representation to learn the contextual prior.

To leverage strong \emph{geometric priors} for indoor environments, we represent the 3D structure for each pixel with a 3D plane equation, rather than raw depth value at each pixel.  By doing so, we take advantage of the fact that indoor environments are often comprised largely of planar surfaces. Since all pixels on the same planar surface have the same plane equation, the 3D structure is piecewise constant in a typical scene, which makes dense predictions of plane equations more robust than alternative representations.

To provide \emph{meaningful supervision} for the network to cover the large solution space, we make use of multiple loss functions that account for both pixel level accuracy (pixel-wise reconstruction loss) and global context consistency (adversarial loss, and scene attribute loss).

The primary contribution of our paper is to propose the task of semantic-structure view extrapolation and present \OURS, a unified framework able to produce a complete room structure and semantic labeling when given a partial observation of a scene. This unified framework is able to handle different camera configurations and input modalities.
The experimental results show that direct prediction of the 3D structure and semantics for the unobserved scene provides a more accurate result than alternative methods.
Both the plane equation encoding and the context model learned from multi-level supervision with large scale indoor scenes help to improve prediction quality.


\section{Related Work}
The general scene understanding problem focuses on understanding what is present in an image, including scene classification \cite{li2010object,xiao2010sun}, semantic segmentation \cite{long2015fully}, depth and normal estimation \cite{garg2016unsupervised,wang2015designing}, etc. In this section, we review prior work on these tasks beyond the visible scene.

\begin{figure*}[t]
    \vspace{-6mm}
    \includegraphics[width=\linewidth]{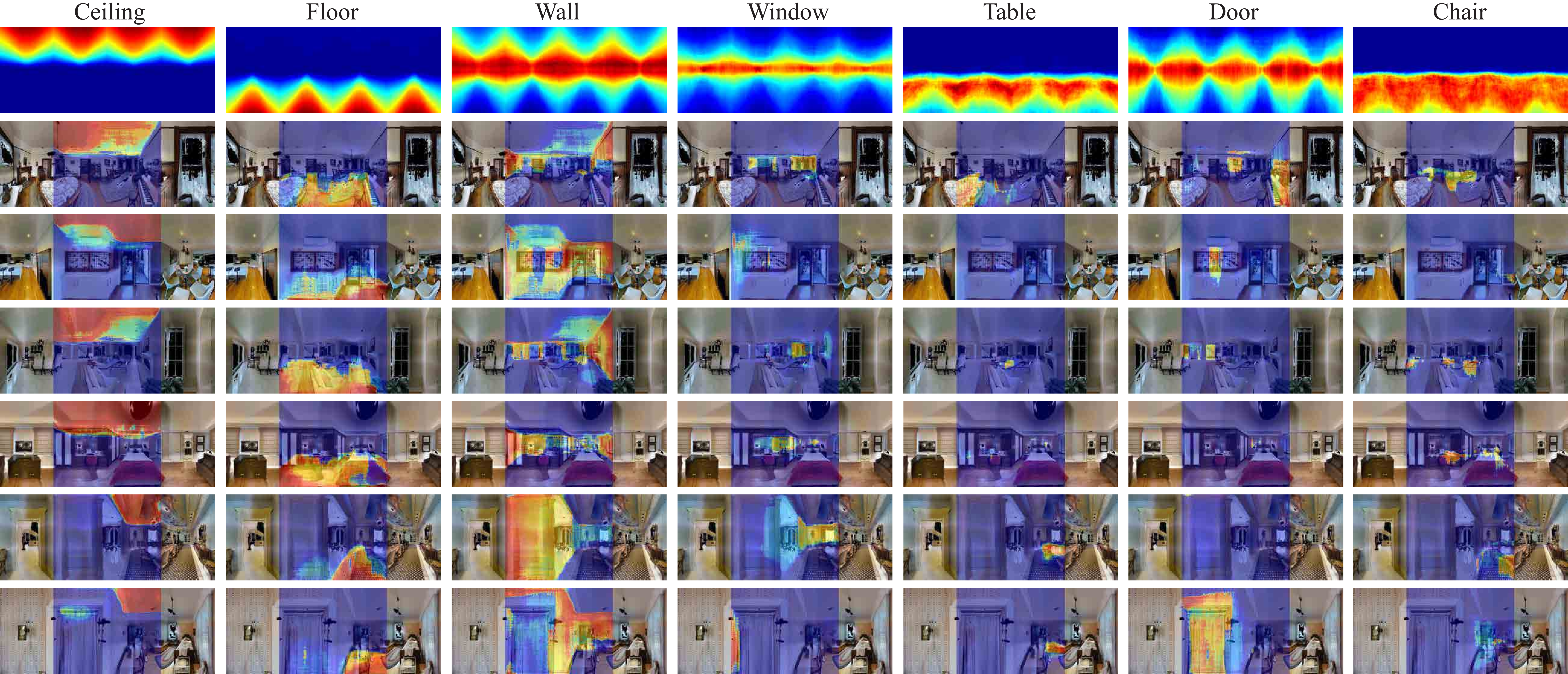}
    \caption{{\bf Probability distribution of semantics.} The first row shows the average distribution of each semantic category over all training examples. The following rows show the predicted probability distribution of semantics from \OURS overlaid  on top of the ground truth testing images. Red areas on the heat maps indicate higher probabilities. }\label{fig:distribution}
    \vspace{-4mm}
\end{figure*}

\mypara{Texture synthesis and image inpainting.}
Texture synthesis methods can be used for image hole filling and image extrapolation \cite{efros1999texture,heeger1995pyramid}.
For example, Barnes et al. \cite{barnes2009patchmatch} fills holes by cloning structures from similar patches. Pathak et. al \cite{pathakCVPR16context} train an autoencoder network.  These methods can achieve very impressive inpainting results for holes in color images. 
However, it is challenging for them to predict image content far outside the field of view, since they don't explicitly model structure and semantics. 

\mypara{Stitching images from the Internet.}
Methods have also been proposed to extrapolate images drastically beyond the field of view using collections of Internet images. For example, Shan et al. \cite{shan2014photo} produce ``uncropped images" by stitching together collections of images captured in the same scene. Hays and Efros \cite{hays2007scene} fill large holes by copying content from similar images in a large collection. While these methods produce impressive results, they only work for scenes where collections are available with many images from nearby viewpoints. 

\mypara{User-guided view extrapolation.}
FrameBreak \cite{zhang2013framebreak} performs dramatic view extrapolation. However, it uses a ``guide image" provided by a person to constrain the image synthesis process. The guide image is chosen from a collection of panorama images, aligned with the input image, and then used to guide a patch-based texture synthesis algorithm.  In this work, we aim to produce an image extrapolation framework that can be used for any common indoor environment without human intervention.

\mypara{Predicting 3D structure in occluded regions.}
Recently there have been many works addressing the problem of shape completion for individual objects \cite{Grasping,rock2015completing,3DShapeNets} or scenes \cite{wadafully,SSCNet,FirmanCVPR2016}. Given a partial observation of an object or scene, the task is to complete the shape of object in the occluded regions within the field of view.  Unlike these methods, \OURS needs to predict the 3D structure outside the field of view, where there is no direct observation, which makes the problem much harder. 


\mypara{Predicting semantic concepts beyond the visible scene.}
Khosla \etal \cite{CVPR14_Khosla} propose a framework to predict the locations of semantic concepts outside the visible scene, e.g., answering questions like ``where can I find a restaurant'' given a street-view image without direct sight of any restaurant.  Although related, their work focuses on outdoor street view scenes and provides only high-level sparse semantic predictions. In contrast, we produce dense pixel-wise predictions for both 3D structure and semantics for pixels outside the observed view for indoor scenes.   



\section{Semantic-Structure View Extrapolation}
We formulate the semantic-structure view extrapolation problem as an image inpainting task by representing both the input observation and output prediction as multi-channel panoramic images. 
The goal of \OURS is to predict the 3D structure and semantics for all missing regions in the input panorama. 
For the semantic prediction, instead of representing it as a discrete category, we model it as a probability distribution over all semantic categories as shown in Fig.\ref{fig:distribution}, which explicitly models the prediction uncertainty. 

\subsection{Whole Room Panoramic Representation}
\label{sec:panoramic-representation}
Traditional view synthesis works \cite{seitz1996view,seitz1995physically}  represent observations and new views using a set of disjoint images with their camera parameters. However this requires the network to handle arbitrary numbers of input views, infer spatial relationships between them, and reason about how scene elements cross image boundaries.  

\begin{figure}[h]

    \includegraphics[width=\linewidth]{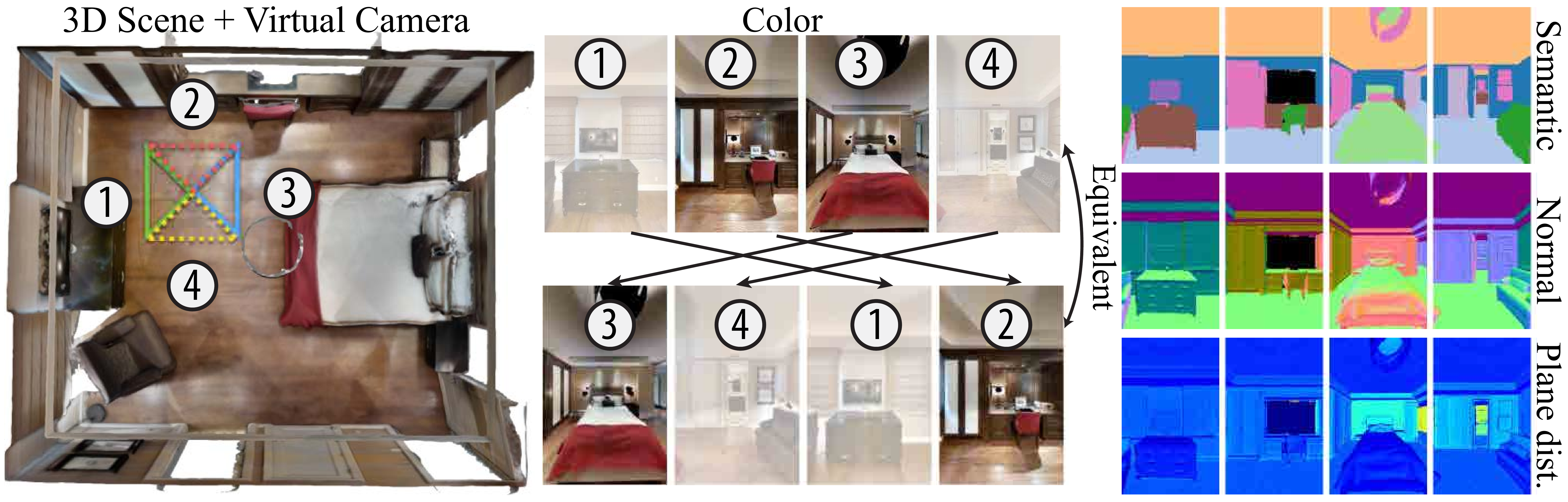}
    \caption{{\bf Whole room representation.} We use a sky-box-like multi-channel panorama to represent 3D scenes. The views are circularly connected, hence, observing the inner two views is equivalent to observing the outer two views of its shifted panorama. }\label{fig:panoview}
    \vspace{-3mm}
\end{figure}

In contrast, we propose to represent the 3D scene using a single panorama where each pixel is labeled with multiple channels of information (color, 3D structure, and semantic) or marked as unobserved. 
This data representation allows the network to learn a consistent whole-room context model by describing both the observed and unobserved parts of the entire scene from a single viewpoint.  It is particularly efficient for deep learning because the observations and predictions are resampled in a regular 2D parameterization suitable for convolution.  Meanwhile, it can naturally support different input camera configurations through reprojection (see Fig.\ref{fig:allcamera}).
Given an observation of a 3D scene reconstructed from registered RGB-D images, we pick a virtual camera center and render the mesh onto four perspective image planes in a sky-box like fashion (see Fig.\ref{fig:panoview}). Each image plane has a $90 \degree$ horizontal FoV and a $116 \degree$ vertical FoV with a image size $256\times160$. Virtual camera centers are chosen depending on the dataset: for the Matterport3D dataset, we use tripod locations; for the SUNCG dataset, we randomly select locations in empty space; for short RGB-D videos, we use the median of all camera centers. 

\subsection{Representing 3D Surfaces with Plane Equations}
While deep networks have been shown to perform well for predicting color pixels and semantic labels, they continue to struggle at predicting high-quality 3D structure. Current methods for direct regressing raw depth values produce blurred results \cite{tateno2017cnn,laina2016deeper,eigen2014depth}, partly due to the viewpoint-dependent nature of depth maps and the large value variance of depth values even for nearby pixels on the same 3D plane.  Surface normal predictions are generally higher quality; however, solving depth from normals is under-constrained and sensitive to noise. Other more complicated encodings, such as HHA \cite{depthRCNN}, are designed for recognition, but cannot be used directly to recover the 3D structure.

In response to these issues, we propose to represent 3D surfaces with their plane equations: surface normal $n$ and plane distance $p$ to the virtual camera origin.
We expect this representation to be easier to predict in indoor environments composed of large planar surfaces because all pixels on the same planar surface share the same plane equation -- i.e., the representation is mostly piecewise constant.  Moreover, the 3D location of each pixel can be solved trivially 
from its plane equation by intersection with a camera ray. 

\begin{figure}[t]
\vspace{-4mm}
    \includegraphics[width=\linewidth]{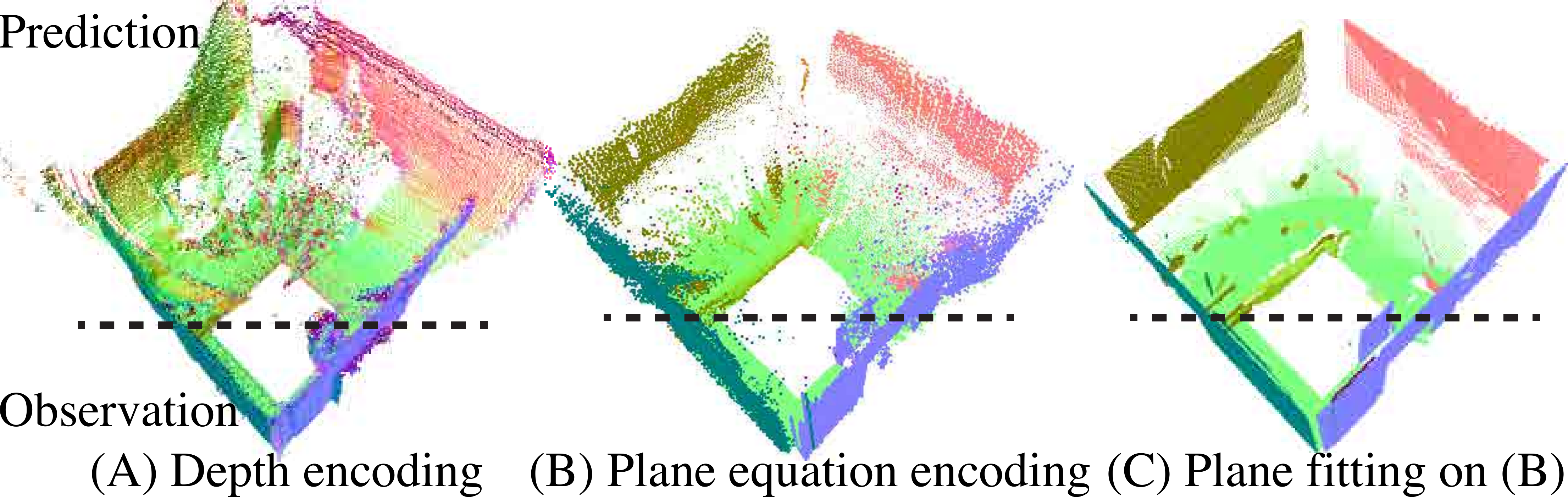}
    \caption{{\bf 3D structure prediction with different encodings.} The plane equation encoding (B) is a better output representation than raw depth encoding (A); its regularization enables the network to predict higher quality geometry. \label{fig:d_pn}}
    \vspace{-3mm}
\end{figure}

Our network is trained to optimize the predicted plane equations. We find this representation of 3D structure to be more effective than raw depth values. Fig.\ref{fig:d_pn} shows the qualitative comparison. We also have a post-processing step to further improve visual quality of the predicted geometry using plane-fitting on the predicted parameters (this step is not included in our quantitative evaluations).


\subsection{Network Architecture}
\begin{figure}[t]
\vspace{-4mm}
    \includegraphics[width=\linewidth]{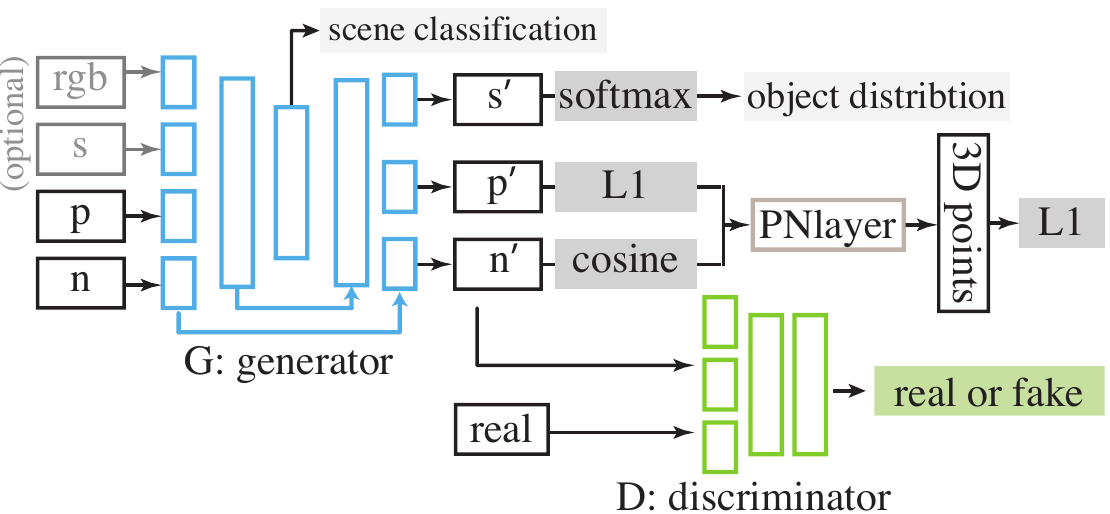}
    \caption{{\bf \OURS network architecture.} the network uses a multi-stream autoencoder structure. A PN-layer is used to ensure consistency between normal and plane distance predictions.}\label{fig:network}
    \vspace{-3mm}
\end{figure}

Our network architecture follows an encoder-decoder structure (Fig.\ref{fig:network}), where the encoder produces a latent vector from an input panorama with missing regions, and the decoder uses that latent vector to produce an output panorama where the missing regions are filled. In this section, we discuss the key features of our network architecture. 

\mypara{Multi-stream network.}
Since our panoramic data representation consists of multiple channels (\eg color, normal, plane distance to the origin, and probability distribution of semantics), we structured our network to process each channel with disjoint streams before merging into and after splitting from the middle layers. In the encoder, each stream is made up of three convolutional layers. The features produced from each stream are merged together by concatenation across channels and then passed through six joint convolutions layers to produce the latent vector. Mirroring this structure, the decoder passes the latent vector through six joint convolutions layers before splitting into multiple streams. This multi-stream structure provides the network a balance of learning both channel-specific parameters within each stream, and joint information through shared layers.



\mypara{Reconstructing 3D surfaces with PN-Layer}
Although our network architecture predicts the parameters of the plane equation as separate channels (surface normals $n$ and plane distances $p$), there is no explicit supervision to enforce the consistency between these two outputs. As a result, we find that with only the individual supervision, the 3D surfaces reconstructed from the predicted parameters tend to be noisy. 
To address this issue, we designed an additional layer in the network (called the PN-Layer) which takes the normal and plane distances as input, and uses the plane equation to produce a dense map of 3D point locations ($x,y,z$) for each pixel based on its respectively predicted $n,p$, and pixel location. This layer is fully differentiable, and therefore an additional regression loss can be added on the predicted 3D point locations in order to enforce the consistency between the surface normal and plane distance predictions. 

\subsection{Network Losses}
When predicting the scene content for the unobserved regions, the plausible solution might not be unique. For example, a valid prediction with slight changes to its locations could still represent an valid solution. To provide the supervision that reflects this flexibility, we use multiple losses to capture three levels of information: pixel-wise accuracy, mid-level contextual consistency using Patch-GAN (adversarial) loss \cite{isola2016image}, and global scene consistency measured by scene category and object distributions. The final loss for each channel is a weighted sum of the three level losses: 


\mypara{Pixel-wise reconstruction loss.}
As part of network supervision, we backpropagate gradients based on the pixel-level reconstruction loss between the prediction and the ground truth panoramas. The loss differs for each output channel. We use softmax loss for semantic segmentation $s$, cosine loss for normal $n$, and $L1$ loss for plane distance $p$ and final 3D point locations $(x,y,z)$.

\mypara{Adversarial loss.}
Following the recent success of generative adversarial networks, we model supervision for generating high-frequency structures in the output panoramas by using a discriminator network \cite{goodfellow2014generative} adapt from PatchGAN \cite{isola2016image}. Similar to the generator, the discriminator network processes each channel with disjoint streams before merging features into shared layers. For the real semantic examples, we converted them into a probabilistic distribution over $C$ classes of size $H \times W \times C$ before feeding them into the discriminator. We adopt the method proposed by Luc \etal \cite{luc:hal-01494296}: For each pixel $i$, given its ground-truth label $l$, we set the probability for that pixel and that label to be $y_{il} = max( \gamma , s(x)_{il})$, where $s(x)_{il}$ is the corresponding prediction from netwotk, and $\gamma=0.8$. For all other classes we set $y_ic=s(x)_ic(1-y_{il})/(1-s(x)_{il})$, so that the label probabilities in $y$ sum to one for each pixel. 

 
\begin{figure}[t]
 \vspace{-4mm}
    \includegraphics[width=\linewidth]{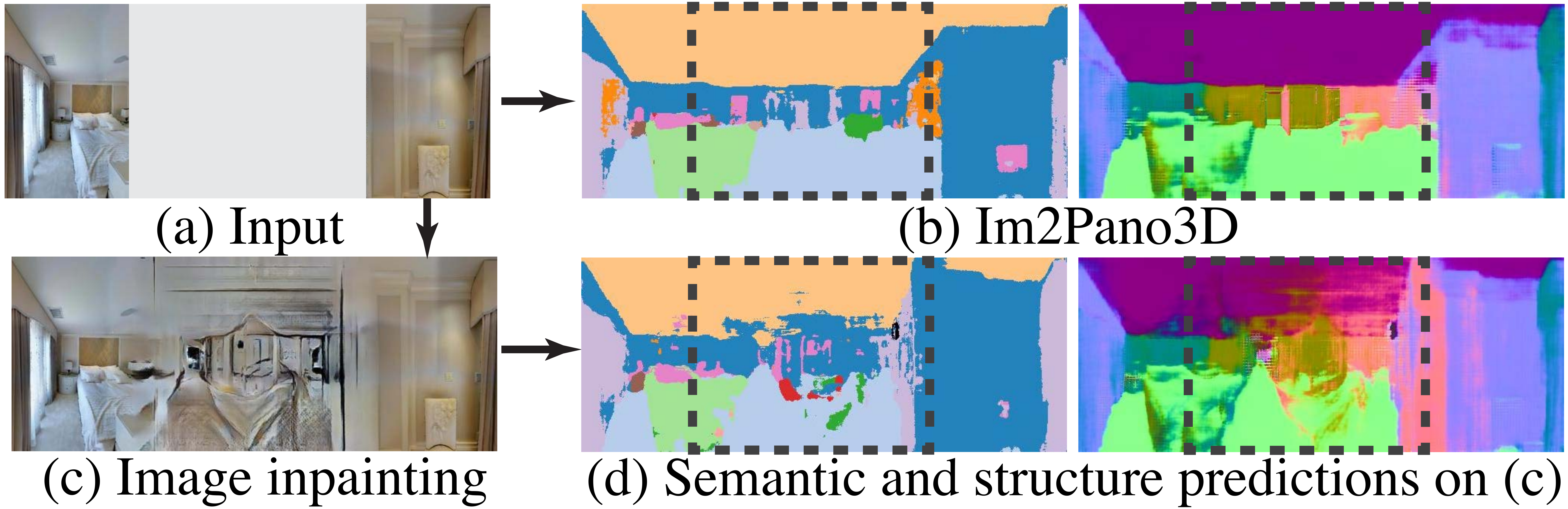}
    \caption{Directly predicting 3D structure and semantics (b) (rgbpn2pns) provides a more accurate result than predicting the same information from generated color pixels (d) (inpaint).  \label{fig:twosteps}}
    \vspace{-3mm}
\end{figure}

\mypara{Scene attribute loss.}
We add additional supervision to the network in order to regularize high level scene attributes such as scene category and overall object distributions. 
To make the network aware of different scene categories, we added two fully connected layers that predict the room category (over 8 scene categories) of the input panorama from its latent code generated by the encoder. We backpropagate gradients directly through the encoder from the softmax classification loss on the scene category predictions.  
Furthermore, we added another auxiliary network that computes the pixel-level distribution of different object classes from its semantic prediction, and backpropagates gradients from comparing this distribution to the ground truth distribution through an $L_1$ loss.
Our ablation studies in Sec.\ref{sec:eval} demonstrate that these additional losses help to improve the semantic predictions, especially for small objects. 


\section{Evaluation}
\label{sec:eval}
\begin{figure*}[t]
 \vspace{-5mm}
    \includegraphics[width=\linewidth]{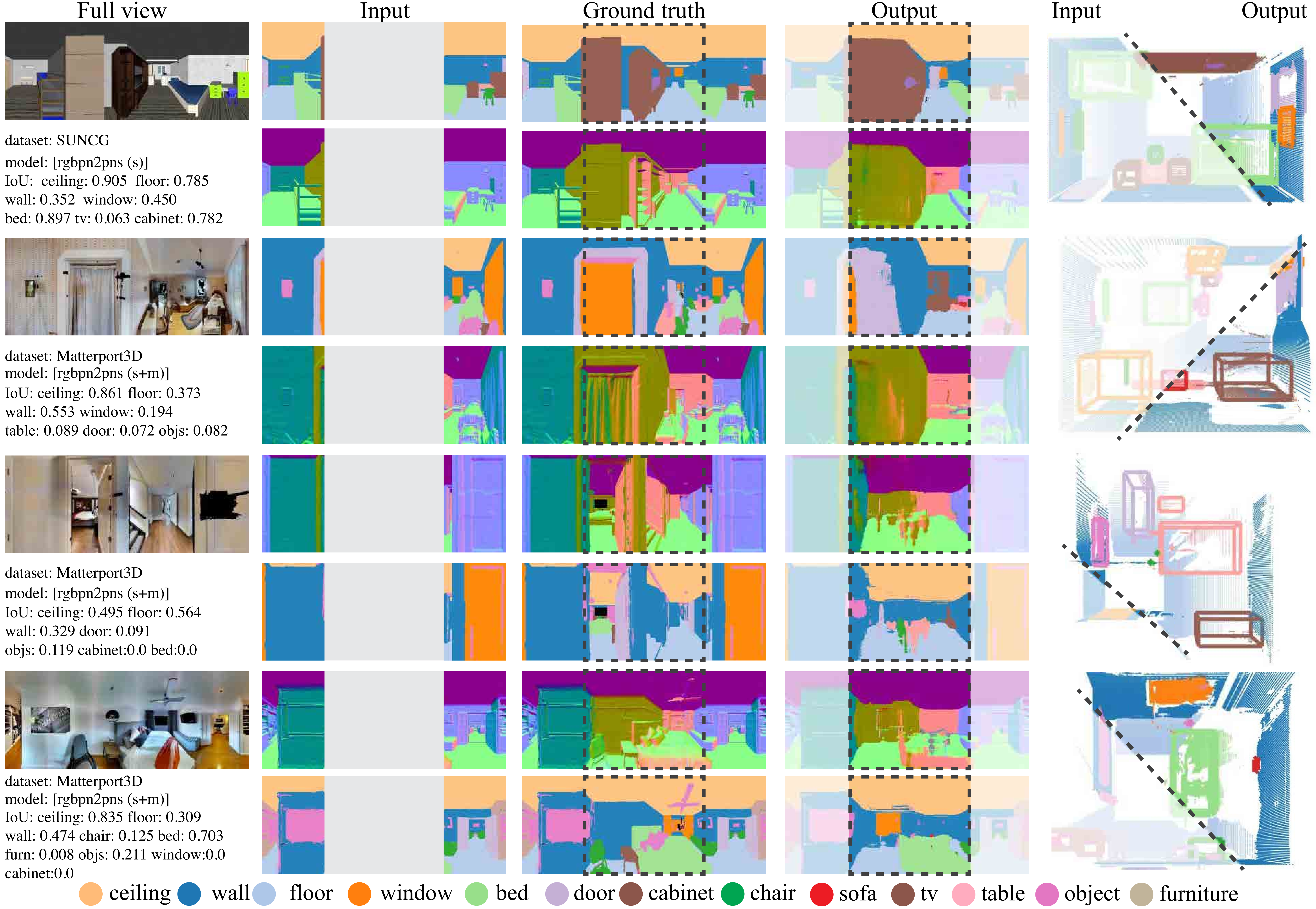}
    \caption{{\bf Qualitative Results.} For each example, we show semantic segmentations labeled using the highest predicted class probability for each pixel, and normal maps from 3D structure predictions. We also show reconstructed 3D point clouds (right column), colored by semantic labels, with bounding boxes around semantically connected components. More results in supplementary material.
    \label{fig:results}}

\end{figure*}

In this section, we present a set of experiments to evaluate \OURS. We not only investigate how well it predicts semantics and structure for unseen parts of a scene, 
but also study the impact of each algorithmic components through ablation studies.
In most of our experiments, we consider the case where the input observation has a $180 \degree$ horizontal and $116 \degree$ vertical FoV, resulting in 50\% partial observation (Fig.\ref{fig:results}). In later experiments, we demonstrate our approach on other camera configurations. All evaluations are performed on unobserved regions only.

\subsection{Datasets}
For our experiments, we use both synthetic (SUNCG \cite{SSCNet}) and real (Matterport3D \cite{Matterport3D}) datasets.  The former is used for pre-training and ablation studies. The latter is used for final evaluation on real data.
\vspace{-1mm}
\begin{itemize}
\setlength{\topsep}{0pt}
\setlength{\parsep}{0pt}
\setlength{\parskip}{0pt}
\setlength{\itemsep}{2pt}

\item[$\Cdot$]{\bf SUNCG \cite{SSCNet}:} This dataset contains synthetically rendered panoramic images with color, depth and semantic of synthetic 3D indoor rooms. In total, we use 58,866 panoramas for training, and 480 for testing.
    
\item[$\Cdot$] {\bf Matterport3D \cite{Matterport3D}:} This dataset contains real RGB-D panoramas captured with a tripod-mounted Matterport camera.  We use color, depth and semantics provided by the dataset, but re-rendered them to form our panoramic representation (Sec. \ref{sec:panoramic-representation}).  In total, we use 5,315 panoramas for training, and 480 for testing.

\end{itemize}
\vspace{-1mm}

\begin{figure}[t]
 \vspace{-4mm}
    \includegraphics[width=\linewidth]{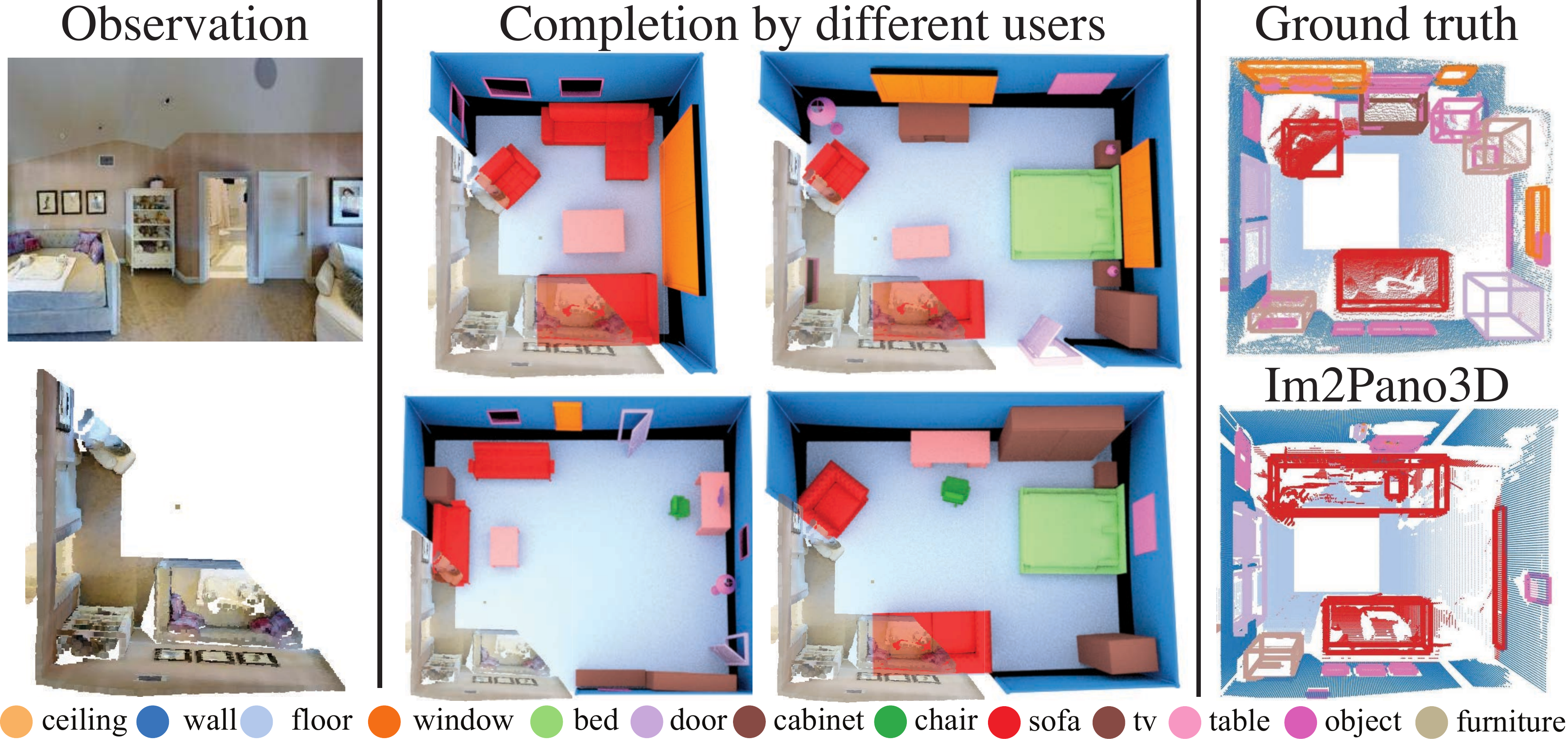}
    \caption{{\bf Human completion.} Left shows the input observations. Middle shows completion results from different users overlaid on the observations. Right shows ground truth and our prediction.\label{fig:human}}
    \vspace{-3mm}
\end{figure}

\begin{table*} [t]
{
\centering
\footnotesize
\setlength{\tabcolsep}{2.5 pt}
\begin{tabular}{l|ccccccc|ccccc|ccccc}
\toprule 
\multicolumn{1}{l|}{models} & \multicolumn{7}{c|}{semantics} & \multicolumn{5}{c|}{3D surface (m)} & \multicolumn{5}{c}{normals ($\degree$)}\tabularnewline
type+loss  & PoG$\uparrow$   & exist$\uparrow$ & size$\downarrow$ & emd$\downarrow$ & IoU$\uparrow$  & acc. $\uparrow$ 	 & incept. $\uparrow$
 & mean $\downarrow$ & med. $\downarrow$ & 0.2{\tiny(\%)} & 0.5{\tiny(\%)} & 1{\tiny(\%)}$\uparrow$ & mean $\downarrow$ & med. $\downarrow$ & 11.25{\tiny(\%)} & 22.5{\tiny(\%)}& 30{\tiny(\%)}$\uparrow$\tabularnewline
\midrule 
pn2pn+A & - & - & - & - & - & - & - & {\bf 0.320} & {\bf 0.119} & {\bf 67.6} & 81.4 & 91 & 38.5 & 5.5 & 70.3 & 74.5 & 76\tabularnewline
d2d+A & - & - & - & - & - & - & - & 0.353 & 0.148 & 63.1 & 79.6 & 90.1 & 59.0 & 41.2 & 12.7 & 29.3 & 38.9\tabularnewline
rgbpns2pns+A+S & {\bf 0.386} & 0.704 & 0.764 & 1.24 & 0.313 & 0.707 & 0.444 & 0.335 & 0.145 & 64.4 & 80.8 & 91.1 & 37.8 & 5.1 & 70.9 & 75.1 & 76.8\tabularnewline
rgbpn2pns+A+S & 0.376 & 0.688 & 0.702 & 1.204 & 0.321 & 0.721 & 0.446 & 0.306 & 0.124 & 67.1 & {\bf 82.4} & {\bf 92.1} & 36.0 & 4.6 & 72.5 & {\bf 76.5} & {\bf 78.2}\tabularnewline
pns2pns+S & 0.379 & 0.613 & {\bf 0.653} & {\bf 1.184} & 0.313 & {\bf 0.728} & 0.375 & 0.416 & 0.227 & 51.8 & 74.3 & 88.9 & {\bf 32.5} & 7.6 & 62.3 & 72.2 & 76.0 \tabularnewline
pns2pns+A & 0.370 & 0.681 & 0.750 & 1.269 & 0.318 & 0.719 & 0.452 & 0.343 & 0.15 & 63.3 & 80.4 & 90.9 & 37.7 & {\bf 4.4} & 72.2 & 76.0 & 77.4\tabularnewline
pns2pns+A+S & 0.382 & {\bf 0.710} & 0.754 & 1.204 & {\bf 0.330} & 0.716 & {\bf 0.463} & 0.339 & 0.151 & 64.0 & 80.8 & 91.1 & 36.9 & 4.6 & {\bf 73.0} & 76.4 & 77.8\tabularnewline
\bottomrule 
\end{tabular}
}
\caption{ Ablation studies on SUNCG. Models are named by their input and output modalities. rgb: color, s: semantic segmentation, d: depth, p: plane distance, n: surface normal. A: adversarial loss, S: scene attribute loss. \label{table:suncg}}
\end{table*}

\begin{table*} [t]
{
\centering
\footnotesize
\setlength{\tabcolsep}{2.4 pt}
\begin{tabular}{ll|ccccccc|ccccc|ccccc}
\toprule
\multicolumn{2}{c|}{models} & \multicolumn{7}{c|}{semantics} & \multicolumn{5}{c|}{3D surface (m)} & \multicolumn{5}{c}{normals ($\degree$)}\tabularnewline
type & train & PoG$\uparrow$   & exist$\uparrow$ & size$\downarrow$ & emd$\downarrow$ & IoU$\uparrow$  & acc. $\uparrow$ 	 & incept. $\uparrow$
 & mean $\downarrow$ & med. $\downarrow$ & 0.2{\tiny(\%)} & 0.5{\tiny(\%)} & 1{\tiny(\%)}$\uparrow$ & mean $\downarrow$ & med. $\downarrow$ & 11.25{\tiny(\%)} & 22.5{\tiny(\%)}& 30{\tiny(\%)}$\uparrow$\tabularnewline
 \midrule 
human & - & 0.303 & 0.650 & 1.474 & 0.943 & 0.203 & 0.522 & - & 0.661 & 0.449 & 29.1 & 57.7 & 78.7 & 49.9 & 17.4 & 51.2 & 58.2 & 60.8\tabularnewline
 \midrule
avg all & m & 0.131 & 0.228 & 1.574 & 2.007 & 0.098 & 0.498 & - & 0.925 & 0.685 & 12.6 & 37.8 & 67.9 & 46.2 & 41.8 & 3.1 & 17.5 & 31.4\tabularnewline
avg type & m & 0.155 & 0.260 & 1.265 & 2.089 &  0.107 &  0.508 & - & 0.905 & 0.668 & 13.8 & 39.6 & 69.6 & 45.8 & 40.4 & 4.5 & 20.7 & 34.0\tabularnewline
nn  & m & 0.126 & 0.531 & 1.901 & 2.820 & 0.078 & 0.302 & - & 1.286 & 0.898 & 15.8 & 33.6 & 56.4 & 65.1 & 58.1 & 23.8 & 31.2 & 34.9\tabularnewline
inpaint & s+m & 0.145 & 0.488 & 1.407 & 1.984 & 0.082 & 0.347 & 0.183 & 0.867 & 0.591 &  19.3 & 46.3 & 72.3 & 59.5 & 50.4 & 23.3 & 32.8 & 37.9 \tabularnewline

rgbpn2pns & s & 0.185 & 0.56 & 1.589 & 1.729 & 0.129 & 0.378 & 0.233 & 0.609 & 0.365 & 32.3 & 63.4 & 82.5 & 47.2 & 20.8 & 43.6 & 54.7 & 59.4\tabularnewline
rgbpn2pns & m & 0.245 & 0.542 & {\bf 0.933 }& 1.535 & 0.174 & {\bf 0.566} & 0.394 & 0.603 & 0.361 & 37.4 & 63.7 & 82.1 & {\bf 39.1 } & 22.4 & 34.9 & 52.6 & 60.4\tabularnewline
rgbpn2pns & s+m & {\bf 0.275} & {\bf 0.616} & 0.936 & {\bf 1.487} & {\bf 0.208} & {\bf 0.566} & {\bf 0.402} & {\bf 0.524} &{\bf  0.280} & {\bf 43.6} & {\bf 69.5 }& {\bf 85.5} & 43.6 & {\bf 19.0} & {\bf 42.9} & {\bf 57.2} & {\bf 62.8} \tabularnewline
\midrule

pns2pns & s & 0.317 & 0.658 & 0.858 & 1.507 & 0.256 & 0.603 & 0.365 & 0.581 & 0.367 & 32.3 & 65.0 & 84.4 & 44.1 & 15.5 & 52.1 & 61.8 & 65.4\tabularnewline
pns2pns & m & 0.304 & 0.618 & {\bf 0.854} & 1.526 & 0.243 & 0.61 & 0.406 & 0.610 & 0.373 & 32.6 & 63.4 & 83.2 & 42.3 & 20.0 & 37.9 & 57.3 & 63.6\tabularnewline
pns2pns & s+m & {\bf 0.355} & {\bf 0.665 }& 0.881 & {\bf 1.425 } & {\bf 0.282} & {\bf 0.623} &{\bf  0.427 } & {\bf 0.563} & {\bf 0.321} & {\bf 38.5} & {\bf 67.6} & {\bf 84.6 } &{\bf  41.2} & {\bf 19.7} & {\bf 40.3} & {\bf 56.9} & {\bf 63.2} \tabularnewline
\bottomrule  
\end{tabular}
}
\caption{Comparing to baseline methods on Matterport3D. Row 2 to 5 shows baseline methods. Our models are named by their input output modalities (same as Tab.\ref{table:suncg}) and training set (s: SUNCG, m: Matterport3D). Bold numbers indicate best performances in each group. \label{table:mp}}
\vspace{-4mm}
\end{table*}

\subsection{Baseline Methods}
To our knowledge, there is currently no prior work that performs our task exactly. To provide baselines for comparison, we consider the following extensions to related work:

\vspace{-2mm}
\begin{itemize}
\setlength{\topsep}{0pt}
\setlength{\parsep}{0pt}
\setlength{\parskip}{0pt}
\setlength{\itemsep}{2pt}
\item[$\Cdot$] {\bf Average distribution (avg)} computes a per pixel average of all images within the training set.
\item[$\Cdot$] {\bf Average distribution by scene category (avg-type)} computes a per pixel average of all training images within the scene category.
The prediction is chosen by the testing images' ground truth scene categories.
\item[$\Cdot$] {\bf Nearest neighbor (nn)} retrieves the nearest neighbor image based on ImageNet features, and uses its semantic segmentation and depth map as the prediction. 
\item[$\Cdot$] {\bf Image inpainting (inpaint)} uses the context encoder of \cite{isola2016image} to directly predict the color pixels in the unobserved regions, followed by a segmentation and plane equation estimation network
with the same architecture as \OURS. Fig.\ref{fig:twosteps} shows an example result.
\item[$\Cdot$] {\bf Human completion (human)} asks people to complete the 3D scene using a 3D design tool \cite{scene-toolkit}, where users can define room layouts and furniture arrangements. Fig.\ref{fig:human} shows a few example completions, and Tab.\ref{table:mp} shows the average performance across four users.
\end{itemize}
\vspace{-1mm}

Tab.\ref{table:suncg} and \ref{table:mp} summarize the quantitative results.  Models are labeled by their input and output modality acronyms; rgb: color, s: semantics, d: depth, p: plane distance, n: surface normal. For example, model [d2d] takes in a depth map as input and predicts the raw depth values of the unobserved regions.
To evaluate the algorithm's performance independent of segmentation accuracy over the observed regions, for the [pns2pns] models, we assume ground truth segmentation for the observed region as input.

\mypara{Evaluation Metrics.}
We measure the quality of the predicted 3D geometric structure with the following metrics:
\vspace{-4mm}
\begin{itemize}
\setlength{\topsep}{0pt}
\setlength{\parsep}{0pt}
\setlength{\parskip}{0pt}
\setlength{\itemsep}{2pt}

\item[$\Cdot$] {\bf Normal angle:} the mean and median angles (in degrees) between prediction and the ground truth, and the percentage of pixels with error less than three thresholds (11.25$\degree$,  22.5$\degree$, 30$\degree$). 

\item[$\Cdot$] {\bf Surface distance:} the mean and median L2 distances (in meters) between final predicted 3D point locations and the ground truth, and the percentage of pixels with error less than three thresholds (0.2m, 9.5m, 1m). 
\end{itemize}
\vspace{-1mm}

We measure the quality of the predicted semantic with the following metrics:
\vspace{-1mm}
\begin{itemize}
\setlength{\topsep}{0pt}
\setlength{\parsep}{0pt}
\setlength{\parskip}{0pt}
\setlength{\itemsep}{2pt}

\item[$\Cdot$]  {\bf Probability over ground truth (PoG):} the pixelwise probability prediction of the ground truth labels averaged within each class then averaged across categories.
\item[$\Cdot$]  {\bf Class existence (exist):} the F1 score of object class existence predictions averaged across all classes (where existence defined as $\geq $ 400 pixels). 
\item[$\Cdot$]  {\bf Class size (size):} the pixel size difference between ground truth and predictions divided by the ground truth size. Evaluated on the object categories with correct existence predictions only.
\item[$\Cdot$]  {\bf Earth Mover's Distance (EMD):}  the average Earth Mover's Distance \cite{rubner1998metric} between the predicted and ground truth 3D points for the categories with correct existence prediction. The weight of each 3D point is assigned with its predicted probability. The probability is normalized to sum up to one for each category. 
We use k-center clustering (k=50) to cluster the 3D points before calculating the EMD.

\item[$\Cdot$] {\bf IoU:} the intersection over union of the most likely predicted pixel label, averaged across all classes.
\item[$\Cdot$] {\bf Accuracy (acc):} the percentage of correctly predicted pixels across all pixels.
\item[$\Cdot$] {\bf Inception score (incept.):} the scene classification score on the generated semantic map using an off-the-shelf image classification network (ResNet50\cite{he2016deep}) trained on ground truth semantic maps, similar to the FCN scores that are normally used to measure the generated image quality \cite{salimans2016improved}. 

\end{itemize}
\vspace{-1mm}
The first four metrics of semantic evaluation are newly introduced for this task. Unlike most semantic segmentation tasks, where predictions are made for pixels directly observed with a camera, our task is to predict semantics for large regions of unobserved pixels, which often contain completely unseen objects.  
For this task, predicting the existence and size of unseen objects is already very difficult and useful for many applications, and thus we include the existence and size metrics, which are invariant to precise object locations. 
We also introduce metrics based on the predicted probability distribution (PoG and EMD), which account for soft errors in position. We use PoG to rank algorithms in our comparisons.

\subsection{Experimental Results}
Tab.\ref{table:suncg} and \ref{table:mp} summarize the quantitative results and Fig.\ref{fig:results} shows qualitative results. More results and visualizations can be found in the supplementary material.

\mypara{Comparing to Baseline Methods.}
Comparing our model [rgbpn2pns (s+m)] to all baseline methods (Tab.\ref{table:mp} row 2-5), our proposed model produces better predictions in terms of both semantics and 3D structure.
In particular, compared to the two-step process of predicting semantic labels over predicted color images in the unobserved regions [inpaint], directly predicting semantic labels in a one-step process can generate a more accurate result (+13\% in PoG and -0.24m in surface distance). Fig.\ref{fig:twosteps} shows a qualitative comparison.

\mypara{Do different surface encodings matter?}
Comparing the model using raw depth values [d2d] to the model using the plane equation encoding [pn2pn] (Tab.\ref{table:suncg} and Fig.\ref{fig:d_pn}), we can see that 
the plane equation encoding provides a strong regularization allowing the network to predict higher quality 3D geometry with lower surface distance  and normal error, 0.03m and 21$\degree$ less respectively.

\mypara{What are the effects of different losses?}
Comparing the model trained with adversarial loss [pns2pns+S+A] and without [pns2pns+S] in Tab.\ref{table:suncg}, we can see that the adversarial loss improves the prediction accuracy for small objects, which is reflected in higher IoU (+2\%). Meanwhile the adversarial loss reduces recall for objects with big pixel area, which is reflected in lower total pixel accuracy (-1.2\%). 
Similarly, the scene attribute loss also improves IoU (+2\%), with a small compromise on total pixel accuracy (-0.3\%).  

\begin{figure}[t]
    \includegraphics[width=\linewidth]{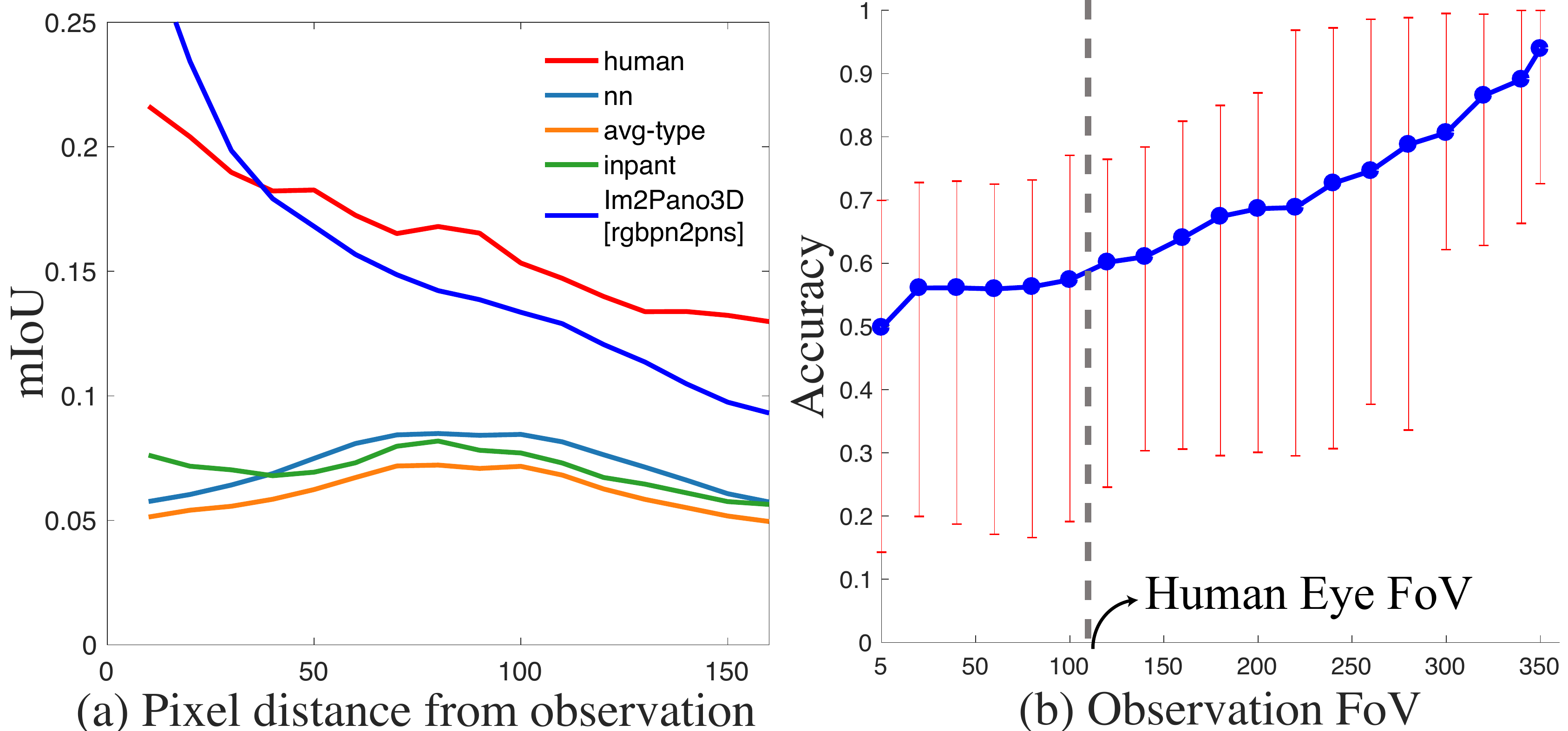}
    \caption{{\bf Experiments.} (a) shows mean IOU with respect to distance from observation. (b) shows accuracy of predictions in the unobserved regions while increasing input horizontal FoV from  $5 \degree$ to $350 \degree$. The error bar shows the error margin across test cases.  \label{fig:plot}}
    \vspace{-3mm}
\end{figure}

\mypara{Does synthetic data help?} 
Comparing our models [pns] and  [rgbpn2pns]  trained with and without the SUNCG dataset and testing on the Matterport3D dataset, we observe that pre-training on SUNCG significantly improves the model's performance, $9\%$ and $4\%$ improvement in PoG respectively. 
In particular, when the input is a segmentation map instead of a color image [pns2pns], the model trained only on SUNCG can even achieve better performance than the model trained on Matterport3D alone (+1.3\% in PoG and -0.08m in surface distance). 
This result demonstrates that training on synthetic data is critical for this task, as it enables the network to learn a rich whole-room contextual prior from a large variety of indoor scenes, which is extremely expensive to obtain with real data.

\mypara{How is accuracy influenced by distance to observation?} 
Fig.\ref{fig:plot} (a) shows the average IoU with respect to its distance to the nearest observed pixel. 
As expected, the performance for \OURS  decreases for pixels that are further from the input observation.
However, the performance is still much higher than other baselines when the region is far from the observation or completely behind the camera, yet still not as high as human performance. 

\mypara{How is accuracy influenced by input FoV?} 
To investigate how the input FoV affects the prediction accuracy, we do the following experiment: we keep the vertical FoV of the input image at $116 \degree$ while steadily increasing the horizontal FoV from  $5 \degree$ to $350 \degree$, and ask the network to predict the structure and semantics for the full panorama. 
Fig.\ref{fig:plot} (b) shows prediction accuracy in the unobserved regions with respect to input FoV, which shows that the prediction accuracy improves as the input FoV increases.

\begin{figure*}[t]
\vspace{-5mm}
\includegraphics[width=\linewidth]{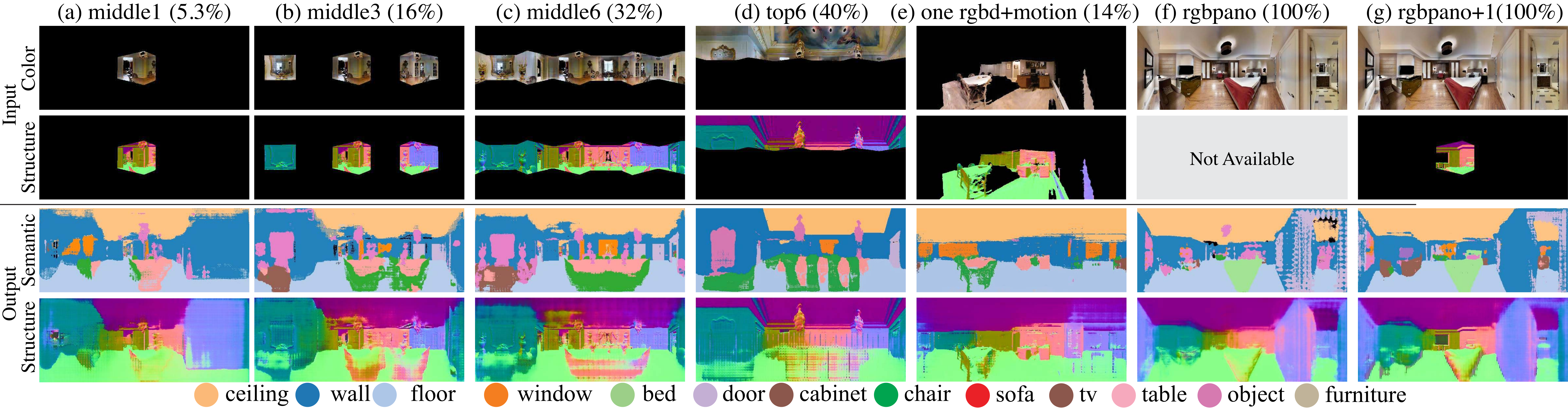}
\caption{{\bf Camera configurations.}  For different camera configurations, \OURS provides a unified framework to efficiently filling in missing 3D structure and semantics of the unobserved scene. The observation coverage is shown in parentheses. 3D structure is represented with normals. The data for [RGB-D+motion] comes from NYUv2 \cite{NYUdataset}. More examples can be found in supplementary material. \label{fig:allcamera} }
\vspace{-3mm}
\end{figure*}

\mypara{Generalizing to different camera configurations.} 
In most of our evaluations, we consider the case where the input observation has a $180 \degree$ horizontal FoV. However, in real robotic applications, systems may be equipped with different types of cameras resulting in different observation FoV patterns. Here we demonstrate how \OURS can generalize to other cases.
The camera configurations we consider includes: single or multiple registered RGB-D cameras such as Matterport cameras (Fig.\ref{fig:allcamera} (a-d)), single RGB-D camera capturing a short video sequence (e), color-only panoramic camera (f), and color panoramic cameras paired with a single depth camera (g).
To improve the ability of the network to generalize to different input observation patterns, we use a random view mask during training. Tab.\ref{table:mask} shows the qualitative evaluation. 
For all of these  camera configurations, \OURS provides a unified framework that effectively fills in the missing 3D structure and semantic information of the unobserved scene. 

\begin{table}
{\footnotesize
\setlength{\tabcolsep}{1.7 pt}
\begin{tabular}{l|ccccccc}
camera & middle1 &middle3 & top6 & bottom6 & middle6 & rgbpano & rgbpano+1\tabularnewline
obs.(\%)& 5.3 & 16.7 & 40.4 & 40.1 & 32.7 & 100 & 100 \tabularnewline
\midrule 
PoG & 0.188 &0.304 & 0.269 & 0.286 & 0.392 & {\bf 0.393} & 0.425\tabularnewline
normal& 29.0 & 13.4 & 14.3 & 14.0 & {\bf 8.8} & 11.3 & 9.5\tabularnewline
surface& 0.454 & 0.238 & 0.237 & 0.322 & {\bf 0.148} & 0.290 & 0.250\tabularnewline
\end{tabular}
}
\caption{{\bf Camera configurations.} The table shows the average PoG, median surface and normal error for each configuration.  Example inputs for each configuration can be found in Fig.\ref{fig:allcamera}.  For models [rgbpano] and  [rgbpano+1], we evaluate on regions that do not have depth observation. For all other models, we evaluate on regions with no color and depth observation. \label{table:mask}}
\vspace{-4mm}
\end{table}

\vspace{-2mm}
\section{Conclusion}
\vspace{-2mm}
We propose the task of semantic-structure view  extrapolation  and  present  \OURS, a unified framework to produce a complete room structure and semantic estimation conditioned on a partial observation of the scene.
Experiments demonstrate that the direct prediction of structure and semantics for the unobserved scene provides more accurate results than alternative approaches. 
However, while \OURS explores the possibilities of whole-room contextual reasoning for 3D scene understanding, the proposed system is still far from perfect. Possible future directions may include: explicitly modeling semantics at the instance-level as opposed to category-level, and exploring alternative data representations that consider occluded regions.

\mypara{Acknowledgment}
This work is supported by Google, Intel, and the NSF(VEC 1539014/ 1539099).  It makes use of data from Matterport3D and Planner5D, and hardware donated by NVIDIA and Intel. Shuran Song is supported by a Facebook Fellowship.  

{\small
\bibliographystyle{ieee}
\bibliography{main}

\begin{thebibliography}{10}\itemsep=-1pt

\bibitem{EMDcode}
http://robotics.stanford.edu/~rubner/emd/default.html.

\bibitem{scene-toolkit}
Scene toolkit: https://github.com/smartscenes/stk.

\bibitem{bar2004visual}
M.~Bar.
\newblock Visual objects in context.
\newblock {\em Nature reviews. Neuroscience}, 5(8):617, 2004.

\bibitem{barnes2009patchmatch}
C.~Barnes, E.~Shechtman, A.~Finkelstein, and D.~B. Goldman.
\newblock Patchmatch: A randomized correspondence algorithm for structural
  image editing.
\newblock {\em ACM Trans. Graph.}, 28(3):24--1, 2009.

\bibitem{borenstein1989real}
J.~Borenstein and Y.~Koren.
\newblock Real-time obstacle avoidance for fast mobile robots.
\newblock {\em IEEE Transactions on Systems, Man, and Cybernetics},
  19(5):1179--1187, 1989.

\bibitem{Matterport3D}
A.~Chang, A.~Dai, T.~Funkhouser, M.~Halber, M.~Niessner, M.~Savva, S.~Song,
  A.~Zeng, and Y.~Zhang.
\newblock {Matterport3D}: Learning from {RGB-D} data in indoor environments.
\newblock 2017.

\bibitem{efros1999texture}
A.~A. Efros and T.~K. Leung.
\newblock Texture synthesis by non-parametric sampling.
\newblock In {\em Computer Vision, 1999. The Proceedings of the Seventh IEEE
  International Conference on}, volume~2, pages 1033--1038. IEEE, 1999.

\bibitem{eigen2014depth}
D.~Eigen, C.~Puhrsch, and R.~Fergus.
\newblock Depth map prediction from a single image using a multi-scale deep
  network.
\newblock In {\em Advances in neural information processing systems}, pages
  2366--2374, 2014.

\bibitem{FirmanCVPR2016}
M.~Firman, O.~Mac~Aodha, S.~Julier, and G.~J. Brostow.
\newblock Structured prediction of unobserved voxels from a single depth image.
\newblock 2016.

\bibitem{garg2016unsupervised}
R.~Garg, G.~Carneiro, and I.~Reid.
\newblock Unsupervised cnn for single view depth estimation: Geometry to the
  rescue.
\newblock In {\em European Conference on Computer Vision}, pages 740--756.
  Springer, 2016.

\bibitem{goodfellow2014generative}
I.~Goodfellow, J.~Pouget-Abadie, M.~Mirza, B.~Xu, D.~Warde-Farley, S.~Ozair,
  A.~Courville, and Y.~Bengio.
\newblock Generative adversarial nets.
\newblock In {\em Advances in neural information processing systems}, pages
  2672--2680, 2014.

\bibitem{depthRCNN}
S.~Gupta, R.~Girshick, P.~Arbelaez, and J.~Malik.
\newblock Learning rich features from {RGB-D} images for object detection and
  segmentation.
\newblock 2014.

\bibitem{hays2007scene}
J.~Hays and A.~A. Efros.
\newblock Scene completion using millions of photographs.
\newblock In {\em ACM Transactions on Graphics (TOG)}, volume~26, page~4. ACM,
  2007.

\bibitem{he2016deep}
K.~He, X.~Zhang, S.~Ren, and J.~Sun.
\newblock Deep residual learning for image recognition.
\newblock In {\em Proceedings of the IEEE conference on computer vision and
  pattern recognition}, pages 770--778, 2016.

\bibitem{heeger1995pyramid}
D.~J. Heeger and J.~R. Bergen.
\newblock Pyramid-based texture analysis/synthesis.
\newblock In {\em Proceedings of the 22nd annual conference on Computer
  graphics and interactive techniques}, pages 229--238. ACM, 1995.

\bibitem{hochberg1978perception}
J.~Hochberg.
\newblock Perception (2nd edn), 1978.

\bibitem{intraub1989wide}
H.~Intraub and M.~Richardson.
\newblock Wide-angle memories of close-up scenes.
\newblock {\em Journal of Experimental Psychology: Learning, Memory, and
  Cognition}, 15(2):179, 1989.

\bibitem{isola2016image}
P.~Isola, J.-Y. Zhu, T.~Zhou, and A.~A. Efros.
\newblock Image-to-image translation with conditional adversarial networks.
\newblock {\em arXiv preprint arXiv:1611.07004}, 2016.

\bibitem{jayaraman2017learning}
D.~Jayaraman and K.~Grauman.
\newblock Learning to look around.
\newblock {\em arXiv preprint arXiv:1709.00507}, 2017.

\bibitem{CVPR14_Khosla}
A.~Khosla, B.~An, J.~J. Lim, and A.~Torralba.
\newblock Looking beyond the visible scene.
\newblock In {\em CVPR}, Ohio, USA, June 2014.

\bibitem{laina2016deeper}
I.~Laina, C.~Rupprecht, V.~Belagiannis, F.~Tombari, and N.~Navab.
\newblock Deeper depth prediction with fully convolutional residual networks.
\newblock In {\em 3D Vision (3DV), 2016 Fourth International Conference on},
  pages 239--248. IEEE, 2016.

\bibitem{li2010object}
L.-J. Li, H.~Su, L.~Fei-Fei, and E.~P. Xing.
\newblock Object bank: A high-level image representation for scene
  classification \& semantic feature sparsification.
\newblock In {\em Advances in neural information processing systems}, pages
  1378--1386, 2010.

\bibitem{long2015fully}
J.~Long, E.~Shelhamer, and T.~Darrell.
\newblock Fully convolutional networks for semantic segmentation.
\newblock In {\em Proceedings of the IEEE Conference on Computer Vision and
  Pattern Recognition}, pages 3431--3440, 2015.

\bibitem{luc:hal-01494296}
P.~Luc, N.~Neverova, C.~Couprie, J.~Verbeek, and Y.~Lecun.
\newblock {Predicting Deeper into the Future of Semantic Segmentation}.
\newblock In {\em {ICCV 2017}}, 2017.

\bibitem{lyle2006importing}
K.~Lyle and M.~Johnson.
\newblock Importing perceived features into false memories.
\newblock {\em Memory}, 14(2):197--213, 2006.

\bibitem{pathakCVPR16context}
D.~Pathak, P.~Kr\"ahenb\"uhl, J.~Donahue, T.~Darrell, and A.~Efros.
\newblock Context encoders: Feature learning by inpainting.
\newblock 2016.

\bibitem{rock2015completing}
J.~Rock, T.~Gupta, J.~Thorsen, J.~Gwak, D.~Shin, and D.~Hoiem.
\newblock Completing {3D} object shape from one depth image.
\newblock 2015.

\bibitem{rubner1998metric}
Y.~Rubner, C.~Tomasi, and L.~J. Guibas.
\newblock A metric for distributions with applications to image databases.
\newblock In {\em Computer Vision, 1998. Sixth International Conference on},
  pages 59--66. IEEE, 1998.

\bibitem{salimans2016improved}
T.~Salimans, I.~Goodfellow, W.~Zaremba, V.~Cheung, A.~Radford, and X.~Chen.
\newblock Improved techniques for training gans.
\newblock In {\em Advances in Neural Information Processing Systems}, pages
  2234--2242, 2016.

\bibitem{seitz1995physically}
S.~M. Seitz and C.~R. Dyer.
\newblock Physically-valid view synthesis by image interpolation.
\newblock In {\em Representation of Visual Scenes, 1995.(In Conjuction with
  ICCV'95), Proceedings IEEE Workshop on}, pages 18--25. IEEE, 1995.

\bibitem{seitz1996view}
S.~M. Seitz and C.~R. Dyer.
\newblock View morphing.
\newblock In {\em Proceedings of the 23rd annual conference on Computer
  graphics and interactive techniques}, pages 21--30. ACM, 1996.

\bibitem{shan2014photo}
Q.~Shan, B.~Curless, Y.~Furukawa, C.~Hernandez, and S.~M. Seitz.
\newblock Photo uncrop.
\newblock In {\em European Conference on Computer Vision}, pages 16--31.
  Springer, 2014.

\bibitem{NYUdataset}
N.~Silberman, D.~Hoiem, P.~Kohli, and R.~Fergus.
\newblock Indoor segmentation and support inference from {RGBD} images.
\newblock 2012.

\bibitem{SSCNet}
S.~Song, F.~Yu, A.~Zeng, A.~Chang, M.~Savva, and T.~Funkhouser.
\newblock Semantic scene completion from a single depth image.
\newblock 2017.

\bibitem{tateno2017cnn}
K.~Tateno, F.~Tombari, I.~Laina, and N.~Navab.
\newblock Cnn-slam: Real-time dense monocular slam with learned depth
  prediction.
\newblock {\em arXiv preprint arXiv:1704.03489}, 2017.

\bibitem{Grasping}
J.~Varley, C.~DeChant, A.~Richardson, A.~Nair, J.~Ruales, and P.~Allen.
\newblock Shape completion enabled robotic grasping.
\newblock 2016.

\bibitem{wadafully}
K.~Wada, K.~Okada, and M.~Inaba.
\newblock Fully convolutional object depth prediction for 3d segmentation from
  2.5 d input.

\bibitem{wang2015designing}
X.~Wang, D.~Fouhey, and A.~Gupta.
\newblock Designing deep networks for surface normal estimation.
\newblock In {\em Proceedings of the IEEE Conference on Computer Vision and
  Pattern Recognition}, pages 539--547, 2015.

\bibitem{3DShapeNets}
Z.~Wu, S.~Song, A.~Khosla, F.~Yu, L.~Zhang, X.~Tang, and J.~Xiao.
\newblock {3D} {ShapeNets}: A deep representation for volumetric shapes.
\newblock 2015.

\bibitem{xiao2010sun}
J.~Xiao, J.~Hays, K.~A. Ehinger, A.~Oliva, and A.~Torralba.
\newblock Sun database: Large-scale scene recognition from abbey to zoo.
\newblock In {\em Computer vision and pattern recognition (CVPR), 2010 IEEE
  conference on}, pages 3485--3492. IEEE, 2010.

\bibitem{zhang2013framebreak}
Y.~Zhang, J.~Xiao, J.~Hays, and P.~Tan.
\newblock Framebreak: Dramatic image extrapolation by guided shift-maps.
\newblock In {\em Proceedings of the IEEE Conference on Computer Vision and
  Pattern Recognition}, pages 1171--1178, 2013.

\bibitem{zhu2017target}
Y.~Zhu, R.~Mottaghi, E.~Kolve, J.~J. Lim, A.~Gupta, L.~Fei-Fei, and A.~Farhadi.
\newblock Target-driven visual navigation in indoor scenes using deep
  reinforcement learning.
\newblock In {\em Robotics and Automation (ICRA), 2017 IEEE International
  Conference on}, pages 3357--3364. IEEE, 2017.

\end{thebibliography}
}

\newpage
\appendix
\section{Additional Algorithm Details}

\paragraph{Network architecture.}
We adapt the encoder-decoder network structure from \cite{pathakCVPR16context,isola2016image}. 
Let C(k,c) denote a Convolution-BatchNorm-ReLU layer with  k filters with c channels, and CD denote a Convolution-BatchNormDropout-ReLU layer with a dropout rate of 50\%.

Encoder: 
[ C(16,x)-C(32,16)-C(64,32) ]x4 - C(192,128) - C(128,256) - C(256,256) - C(256,256) - C(256,256) - C(512,256)

Decoder: CD(512,256) - CD(512,256) - CD(512,256) - C(512,256) -C(512,128) - C(256,128) - [C(256,64) - C(128,64) - C(64,x)]x3 

Discriminator: [C(x,16)]x3 - C(16,64) - C(64,128) - C(64,512).

The number of channels ($x$) of the encoder's first layer and the decoder's last layer depends on the stream's mortality: for color stream $x=3$, for normal stream $x=3$, for plan distance stream $x=1$,  for segmentation stream  $x=1$ in encoder, and $x=13$ in decoder and discriminator.

\paragraph{Training details.}
We implement our network architecture in Torch7.
We randomly initialize all layers by drawing weights from a Gaussian distribution with mean 0 and standard deviation 1.  
For the final model, we pretrain the network on SUNCG dataset for 20 epochs, and finetune on Matterport3D for 20 epochs,both with batch size 3.

\paragraph{Combined loss.}
The following equations describe the details on how we weight different losses for each channel, note that we add the loss from PN-layer only after 1000 iteration to avoid an unstable gradient:

The combined loss for plane distance: 
$ L_{plane} = \lambda_1L_1  + \lambda_2L_{adv} + \lambda_3L_{pn} $
where $\lambda_1 = 0.4, \lambda_2 =  0.01 $, and $\lambda_3 =0.001$ when training iteration $>1000$ $\lambda_3 = 0$ otherwise. \\

The combined loss for surface normal:
$ L_{normal} = \lambda_1L_{cosine}  + \lambda_2L_{adv} + \lambda_3L_{pn}$
,where $\lambda_{cos} = 0.4, \lambda_2 =  0.01 $, and $\lambda_3 =0.001$ when training iteration $>1000$ $\lambda_3 = 0$ otherwise. \\

The combined loss for semantics:
$ L_{semantic} = \lambda_1L_{softmax}  + \lambda_2L_{adv} + \lambda_3L_{distribution}$, where $\lambda_{cos} = 0.4, \lambda_2 =  0.01 $, and $\lambda_3 =0.01$. 

\paragraph{PN-Layer.}
The PN-Layer takes in a predicted normal map and plane distance map and calculates the final 3D point location for each pixel. If the pixel normal prediction is $\vec{n} =  (n_x,n_y,n_z)$ (normalized to be unit length), the plane distance prediction is $p$, the camera intrinsics matrix is $K = [f_x,0,c_x;0,f_y,c_y;0,0,1] $,  the virtual camera center is at $\vec{P_0}$, and the 2D pixel location is $(x_i,y_i, 1)$, then the computed 3D point location $\vec{P} = (x,y,z)$ is 
$$\vec{v} = (\frac{x_i-c_x}{f_x},\frac{y_i-c_y}{f_y}, 1)$$
$$\vec{P} = \vec{P_0}-\frac{\vec{P_0} \cdot \vec{n} + p}{\vec{v} \cdot  \vec{n} } \vec{v}$$
When $\vec{P_0}$ is the origin, we can simplify the equation to:
$$\vec{P} =-\frac{p}{\vec{v} \cdot  \vec{n} } \vec{v}$$

\section{Details on evaluation metrics}
In this section we provide details on the new evaluation metrics: PoG, size, and EMD.  We consider only pixels in the unobserved regions of the panorama for all evaluations.

\paragraph{Probability over ground truth (PoG) } Let $p_{xc}$ be the predicted probability of pixel $x$ for class $c$, where  $ 0 \leq p_{xc} \leq 1$ and $\sum_{c=1}^{13}{p_{xc}} =1 $. Let $G_c$ be the collections of all pixels in the ground truth segmentation map with label equals to class $c$ and $n(G_c)$ be the total number of pixels in this collection. Then $PoG$ for class $c$ is defined as follows:
$$PoG_c = \dfrac{\sum_{i\in G_c}{p_{xc}}}{n(G_c)}$$

\paragraph{Class Size:}  Let $n(G_c)$ be the total number of pixels in the ground truth segmentation map with labels equal to class $c$ and $n(P_c)$ be the total number of pixels in predicted segmentation map equals to class $c$. For each object class $c$ where $n(G_c) \geq 400$ and $n(G_c) \geq 400$,  the size difference $Size_c$ is defined as follow:
$$Size_c = \dfrac{|n(G_c)-n(P_c)|}{n(G_c)} $$

\paragraph{Earth Mover's Distance (EMD): }  Computing the EMD is expensive for two arbitrary 2D distributions.   So, we approximate the calculation using the implementation from  \cite{EMDcode}.   We first cluster all the pixels in $G_c$ into 50 clusters using the K-center algorithm, $G_c = \{(g_1,w_{g1}), (g_2,w_{g2}),...,(g_{50},w_{g50}) \}$, where $g_1$ is the cluster representative (cluster center) and $w_{gi}$ is the weight of the cluster $w_{gi} = n(G_{ci})/n(G_{c})$ for each cluster. 
We also cluster all pixels in the predicted probability distribution map with $p_{xc} > 0.01$  into 50 clusters, 
$Q_c = \{(q_1,w_{q1}), (q_2,w_{q2}),...,(q_{50},w_{q50}) \}$, where $q_i$ is the representative of cluster $ Q_{ci} $ (cluster center) and $w_{qi}$ is the weight of the cluster $w_{pi} = \sum_{x \in Q_{ci}}p_{xc}\sum_{x \in Q_{c}}{p_{xc}}$ for each cluster. 
Then we find the flow $F=[f_{ij}]$ among all flows $f_{ij}$ between $g_{i}$ and $q_{j}$ that minimizes the overall cost: $$\min {\sum _{i=1}^{m}\sum _{j=1}^{n}f_{ij}d_{ij}}$$
subjected to the constraints:
$$f_{i,j}\geq 0,1\leq i\leq m,1\leq j\leq n$$
$$\sum _{j=1}^{n}{f_{ij}}\leq w_{gi},1\leq i\leq m $$
$$\sum _{i=1}^{m}{f_{ij}}\leq w_{qj},1\leq j\leq n $$
$$\sum _{i=1}^{m}\sum _{j=1}^{n}f_{i,j}=\min \{ \sum _{i=1}^{m}w_{gi},\sum _{j=1}^{n}w_{qj} \}$$
The earth mover's distance is defined as the work normalized by the total flow:
$$ EMD_c(G_c,Q_c)={\frac {\sum _{i=1}^{m}\sum _{j=1}^{n}f_{ij}d_{ij}}{\sum _{i=1}^{m}\sum _{j=1}^{n}f_{ij}}}$$

\section{Additional Experiment Results}

\begin{figure}[h]
    \includegraphics[width=\linewidth]{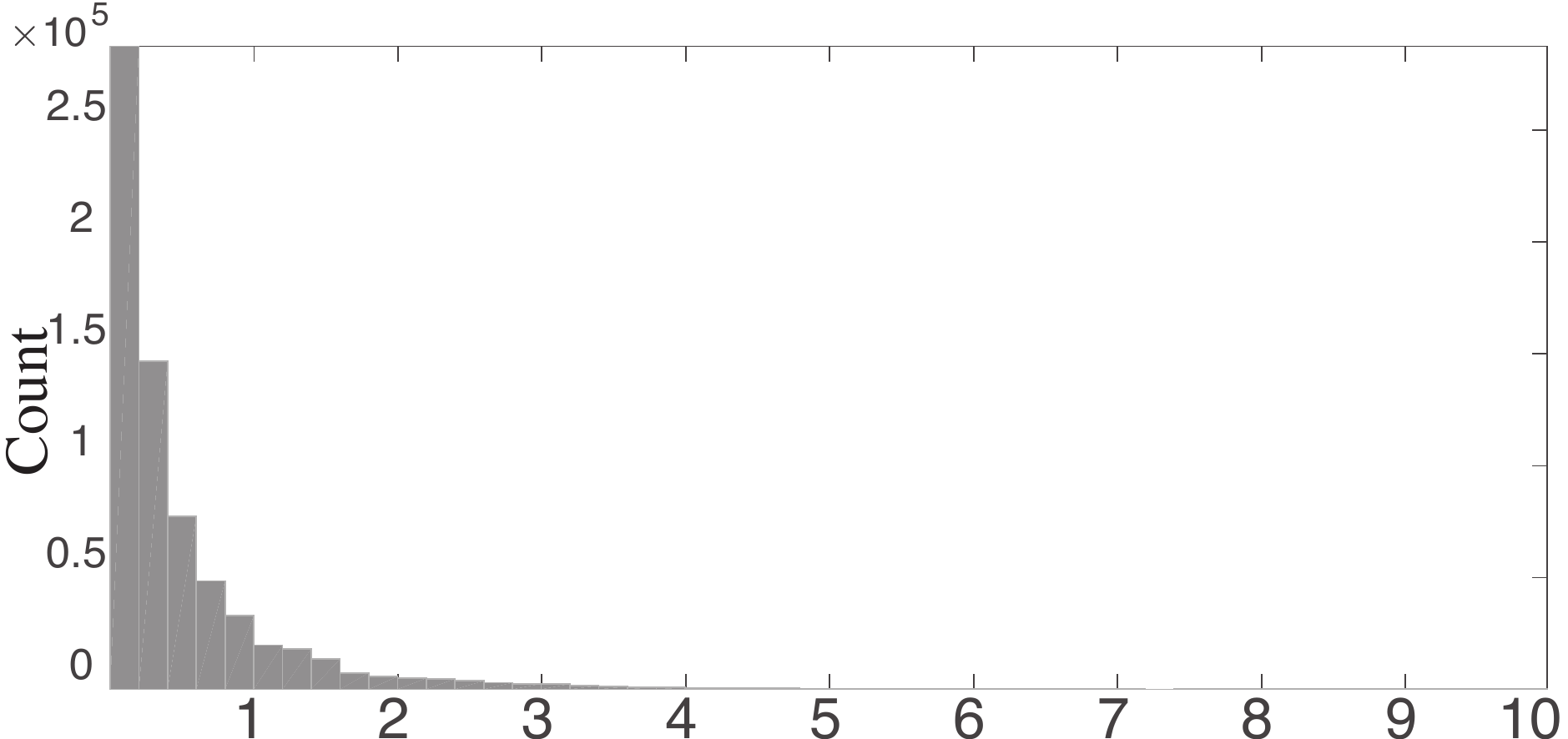}
    \caption{3D surface Error distribution (in meters).\label{fig:errordepth}}
    \vspace{-4mm}
\end{figure}

\begin{figure}[h]
    \includegraphics[width=\linewidth]{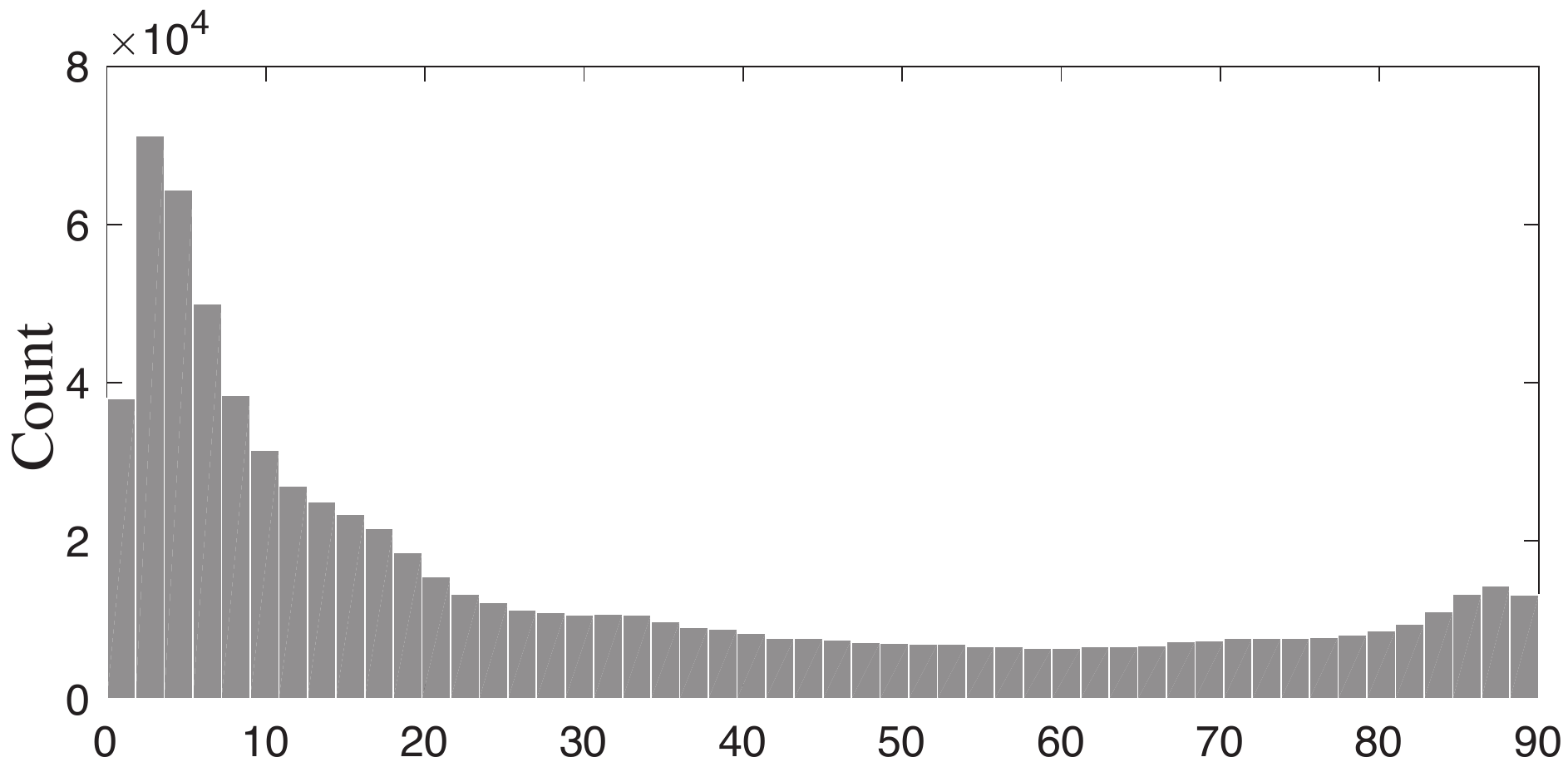}
    \caption{Normal Error distribution (in degree).\label{fig:errornormal}}
        \vspace{-4mm}
\end{figure}

\paragraph{Normal and depth error distribution}
Fig.\ref{fig:errordepth} shows a histogram of the 3D surface distance error (L2 distance in meters) on the test set. We can see that most of the errors are within 1 meter to the ground truth.
Fig. \ref{fig:errornormal} shows a histogram of the angular error in predicted surface normals on the test set. We can see that most of the normal predictions fall within 0\degree to 20\degree of ground truth.

\begin{figure}[h]
    \includegraphics[width=\linewidth]{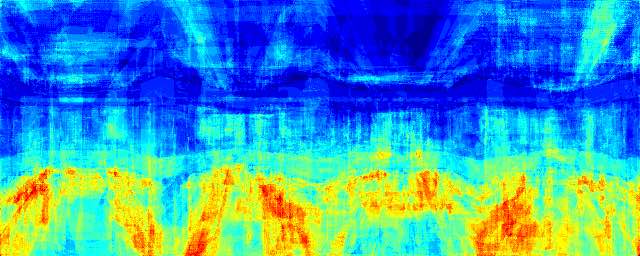}
    \caption{ Error distribution map. red indicates higher error\label{fig:errormap}}
        \vspace{-4mm}
\end{figure}

\paragraph{Where should the camera look?} 
As shown in the main paper, observing more of the scene typically leads to higher accuracy in predicting the rest of the scene. However, we also notice that the observation pattern (defined by the cameras' locations and their viewing angles) also has a strong impact on prediction accuracy. Comparing [top6] and [middle3] in Tab.3 of the main paper, we can see that although [top6] has more cameras and view coverage, its prediction accuracy is lower because the  cameras are looking at regions of scene with low information density (\eg ceilings). Fig.\ref{fig:errormap} shows the spatial distribution of semantic prediction error. The red regions (indicating areas with higher error) in the lower half of the panorama show that some parts of the scene are harder to predict than others. This error map could help determine camera placement on a domestic robot, where the camera should be oriented to look at those regions in order to reduce overall uncertainty.

\begin{figure*}[t]

\includegraphics[width=\linewidth]{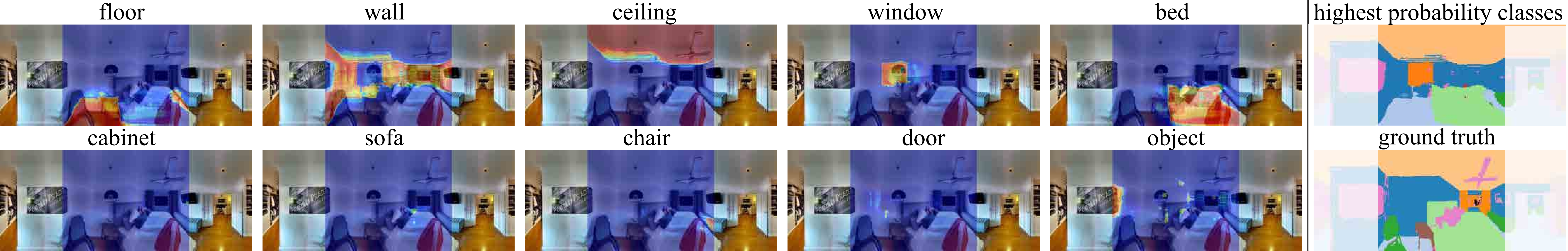}
\caption{In this example, the input RGB-D observation contains a view of a room with a tv on the right wall and a picture (partially) on the left wall. In the ouput the network not only complete the unobserved part of the painting and the chair but also correctly predict the location of a bed . The network also correctly predicts the existence of a window, however, with a different location compared to ground truth. On the other hand, the prediction misses several objects such as cabinets, and pillows. 
}\label{fig:de_1}\vspace{4.5mm}

\includegraphics[width=\linewidth]{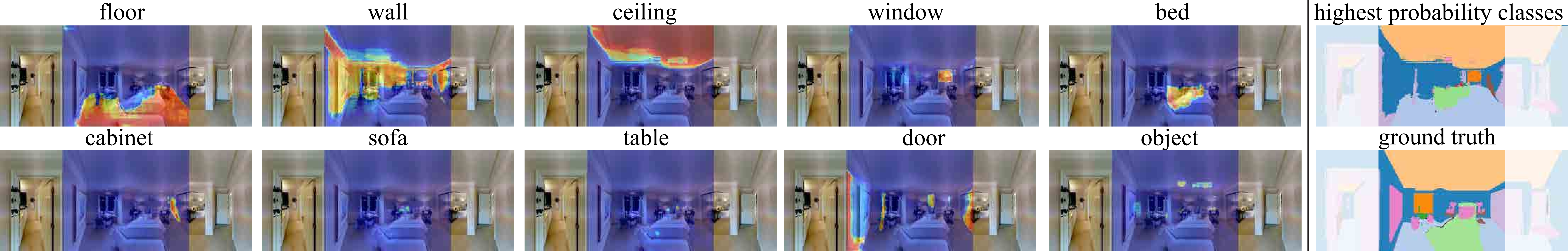}
\caption{In this example, the input RGB-D observation contains a view of a bathroom (through a doorway), and half of a closet. These elements typically co-exist in a bedroom. As a result the network predicts the scene category to be bedroom. In particular, the network predicts the semantics and 3D structure of a bed, a window, and a cabinet in the missing region, without any direct observation of these objects from the input. While the network correctly predicts the existence of these objects, and makes a reasonable prediction of the room layout, we can see that the predicted bed is smaller than that of the ground truth. Also, the predicted window and cabinets are in different locations compared to the ground truth. From the probability distribution maps, we can also observe that the network has several hypotheses for the potential locations of doors, but with lower probability - indicating the uncertainty of the predictions.
}\label{fig:de_22}\vspace{4.5mm}
\includegraphics[width=\linewidth]{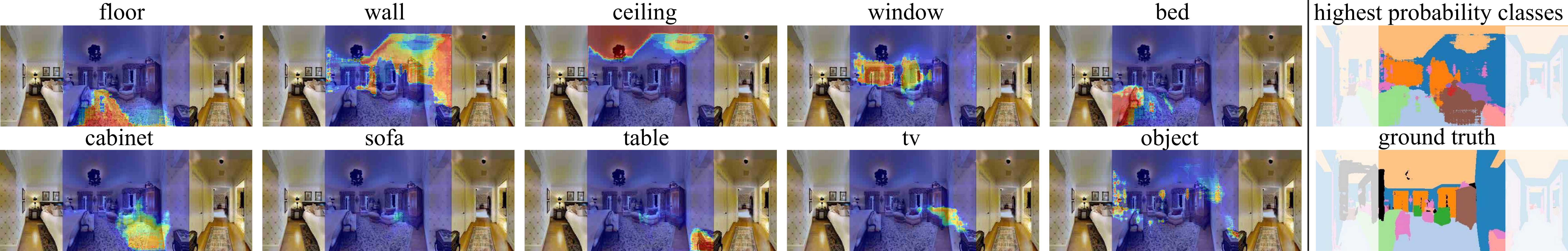}
\caption{In this example, the input RGB-D observation contains a partial view of a bed and a door way. As a result, the network successfully completes the 3D structure of the bed and other room structures within the missing region. The network also predicts the existence and rough locations of windows without directly observing them in the input. Furthermore, the network predicts the existence of a TV and cabinet at reasonable locations (across the bed) based on its learned contextual prior of bedrooms, which are very likely to have TVs placed on top of cabinets. While the prediction of TV and cabinet is plausible, it is different from the ground truth.
}\label{fig:de_4} \vspace{4.5mm}
\includegraphics[width=\linewidth]{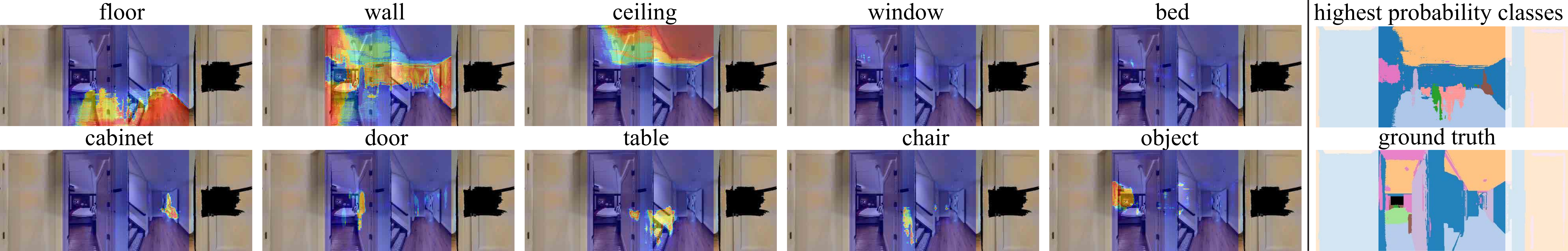}
\caption{In this example, the input RGB-D observation contains only a view of a white wall and a white door. The network completes the scene as if it were a dining room with a table and chairs surrounding it. Although the completed scene looks plausible, it is very different from ground truth - which is a hallway with a partial view of a bedroom (through a doorway). This example demonstrates cases where the partial input observation does not contain sufficient information for the network to perform a prediction close to the ground truth.}
\label{fig:de_3}

\end{figure*}

\begin{figure*}[h]
    \includegraphics[width=\linewidth]{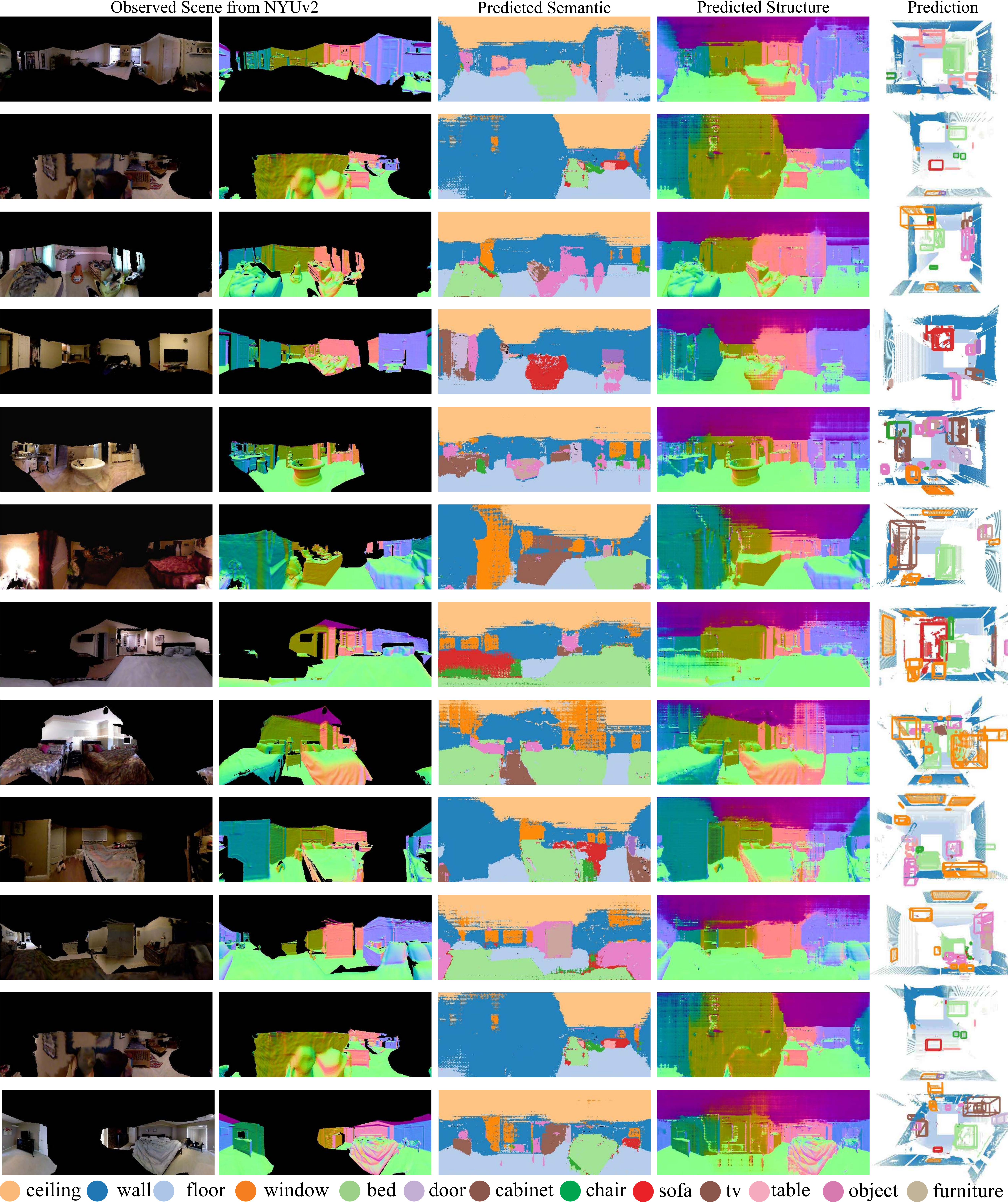}
    \caption{Short sequence observation from NYU dataset \cite{NYUdataset}}\label{fig:nyu_seq}
\end{figure*}

\begin{figure*}[t]
\vspace{-5mm}
{\centering
\includegraphics[width=0.97\linewidth]{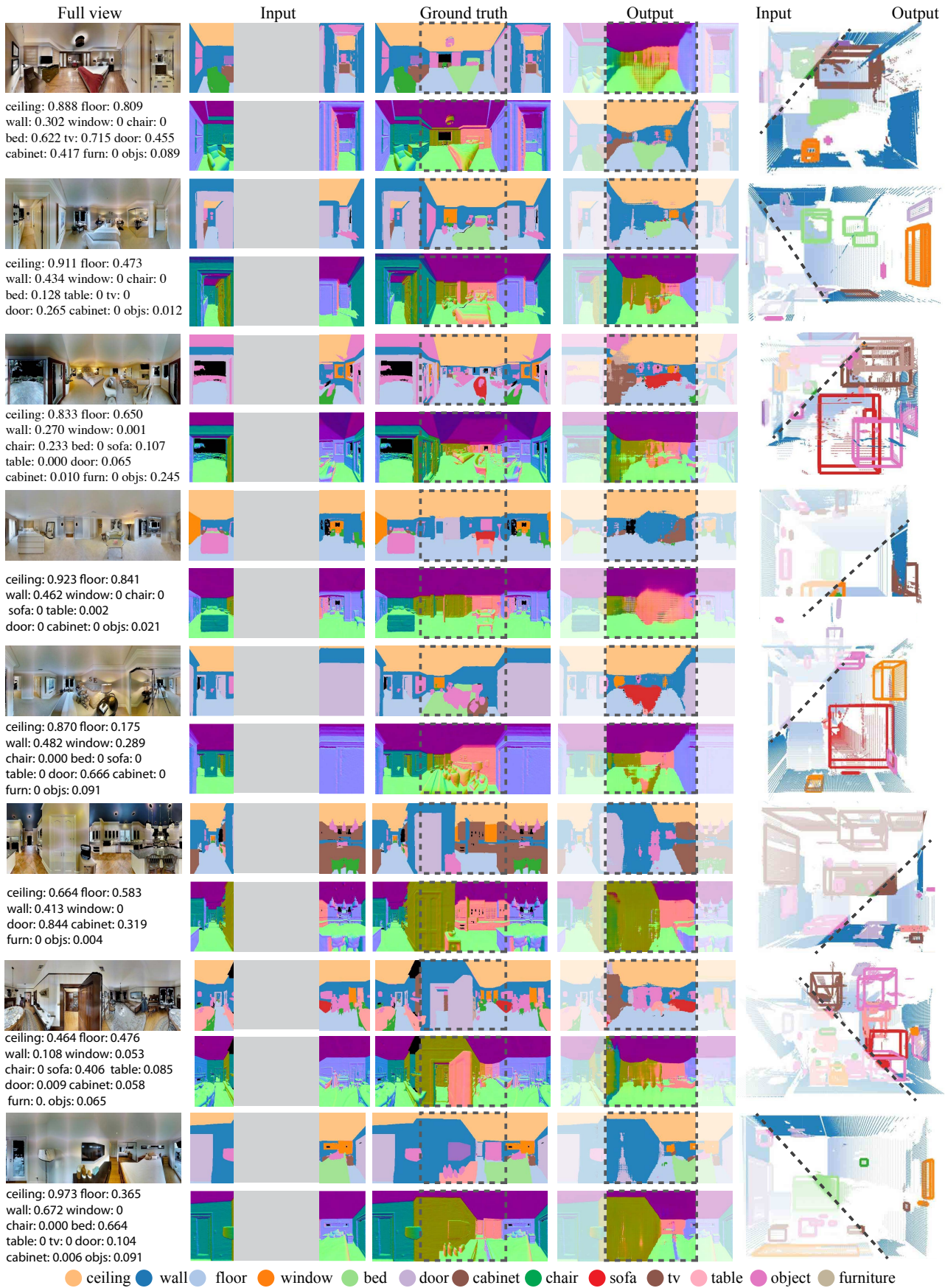}
}
\caption{Additional result on Matterport3D \cite{Matterport3D} dataset.  Model: rgbpn2pns (s+m)}
 \label{fig:mp_result}
\end{figure*}

\paragraph{Additional Quantitative results}
Figs \ref{fig:de_1} to \ref{fig:de_3} show some typical prediction results with detailed analysis.
Fig.\ref{fig:nyu_seq} shows additional result for short video sequence input from NYU dataset \cite{NYUdataset}.
Fig.\ref{fig:mp_result} shows additional result  on  SUNCG dataset \cite{SSCNet}. 
Fig.\ref{fig:mp_result} shows additional result  on  Matterport3D dataset \cite{Matterport3D}.

\paragraph{Per-category performance breakdown}
Tables \ref{table:pog} and \ref{table:size} show the per-category performance for model rgbpn2pns and pns2pns on the Matterport3D dataset evaluated with PoG, IoU, EMD,  class existence, and class size. We find that the network performs well on predicting room structure (\ie wall, floor, ceiling) and large furniture (\eg bed, door, etc.). Although the network finds it challenging to predict smaller objects in precisely the same locations as ground truth (as expected), it still performs well at predicting their existence.

\begin{figure*}[t]
    \includegraphics[width=\linewidth]{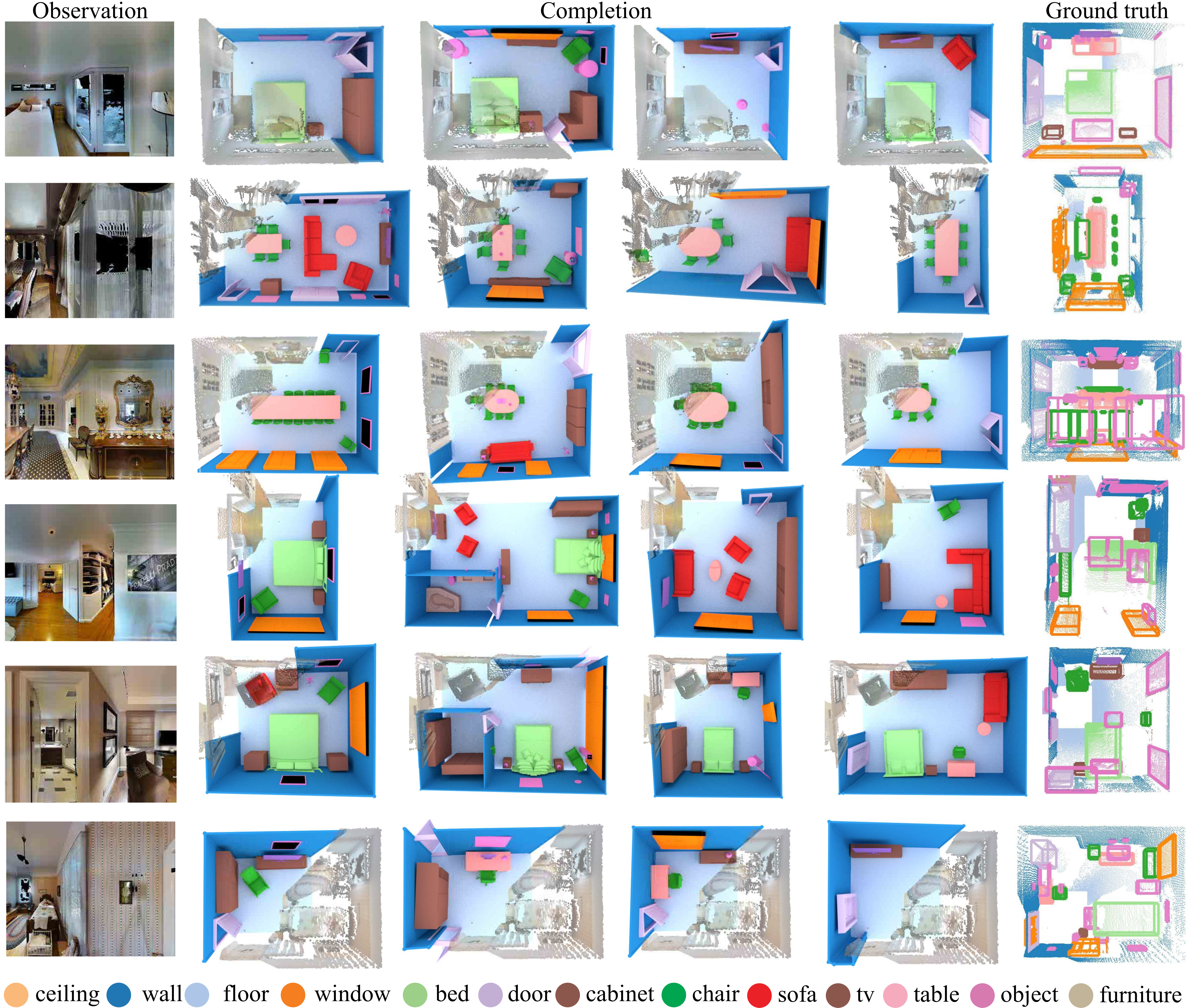}
    \caption{{\bf Example human completions}}\label{fig:human_more}
\end{figure*}

\paragraph{Example human completions}
Fig.\ref{fig:human_more} shows more examples of human completion. In some examples, completion results from different users can be quite consistent (e.g. row 4), while in other examples, different users can generate very different completion results. For example, in row 3, two users design their completion predictions thinking that the room is a bedroom, while the other two design their predictions thinking that the room is a living room. Interestingly, regardless of which room type with which the user formulates his/her completion result, all predictions include the existence of windows on walls in an arrangement consistent with ground truth

\begin{table*}
\centering 
\setlength{\tabcolsep}{3.5 pt}
\begin{tabular}{ccccccccccccccc}

\hline 
 & ceiling & floor & wall & window & chair & bed & sofa & table & tv & door & cabinet & furn & objs & mean\tabularnewline
\hline 
human & 0.874 & 0.624 & 0.593 & 0.103 & 0.137 & 0.331 & 0.446 & 0.293 & 0.200 & 0.074 & 0.248 & 0.001 & 0.032 & 0.305\tabularnewline
rgbpn2pns & 0.814 & 0.661 & 0.561 & 0.183 & 0.156 & 0.268 & 0.149 & 0.138 & 0.065 & 0.222 & 0.145 & 0.060 & 0.154 & 0.275\tabularnewline
pns2pns & 0.797 & 0.622 & 0.651 & 0.287 & 0.159 & 0.361 & 0.282 & 0.235 & 0.130 & 0.334 & 0.286 & 0.225 & 0.242 & 0.355\tabularnewline
\hline 
\end{tabular}
 \vspace{1mm}
\caption{Probability over Groundtruth (higher is better) \label{table:pog} } \vspace{4mm}

\centering 
\begin{tabular}{ccccccccccccccc}
\hline 
 & ceiling & floor & wall & window & chair & bed & sofa & table & tv & door & cabinet & furn & objs & mean\tabularnewline
\hline 
human & 0.703 & 0.424 & 0.348 & 0.079 & 0.112 & 0.280 & 0.233 & 0.207 & 0.091 & 0.048 & 0.112 & 0.001 & 0.025 & 0.205\tabularnewline
rgbpn2pns & 0.704 & 0.505 & 0.404 & 0.143 & 0.096 & 0.204 & 0.12 & 0.103 & 0.044 & 0.139 & 0.13 & 0.034 & 0.073 & 0.208\tabularnewline
pns2pns & 0.729 & 0.511 & 0.475 & 0.241 & 0.134 & 0.285 & 0.231 & 0.16 & 0.091 & 0.242 & 0.246 & 0.149 & 0.173 & 0.282\tabularnewline
\hline 
\end{tabular}
 \vspace{1mm}
\caption{Intersection over union  (higher is better). }\vspace{4mm}
\setlength{\tabcolsep}{3.4 pt}
\centering 
\begin{tabular}{ccccccccccccccc}
\hline 
 & ceiling & floor & wall & window & chair & bed & sofa & table & tv & door & cabinet & furn & objs & mean \tabularnewline
\hline 
human & 0.504 & 0.398 & 0.770 & 1.593 & 0.76 & 0.414 & 1.630 & 0.485 & 0.329 & 1.59 & 1.704 & 0.968 & 1.144 & 0.946\tabularnewline
rgbpn2pns & 0.406 & 0.487 & 0.684 & 2.263 & 1.639 & 1.434 & 2.153 & 1.551 & 1.943 & 1.597 & 2.009 & 2.194 & 0.972 & 1.487\tabularnewline
pns2pns & 0.409 & 0.527 & 0.662 & 2.119 & 1.548 & 1.434 & 2.039 & 1.455 & 2.044 & 1.473 & 1.878 & 2.059 & 0.875 & 1.425\tabularnewline
\hline 
\end{tabular}
 \vspace{1mm}
\caption{Earth mover distance  (lower is better).}\vspace{4mm}

\centering

\begin{tabular}{ccccccccccccccc}
\hline 
 & ceiling & floor & wall & window & chair & bed & sofa & table & tv & door & cabinet & furn & objs & mean\tabularnewline
\hline 
human & 0.995 & 0.995 & 1 & 0.687 & 0.694 & 0.672 & 0.523 & 0.654 & 0.250 & 0.535 & 0.691 & 0.063 & 0.656 & 0.647\tabularnewline
rgbpn2pns & 0.995 & 0.986 & 1 & 0.525 & 0.607 & 0.548 & 0.407 & 0.491 & 0.125 & 0.645 & 0.512 & 0.315 & 0.857 & 0.616\tabularnewline
pns2pns & 0.996 & 0.979 & 1 & 0.592 & 0.555 & 0.592 & 0.563 & 0.561 & 0.294 & 0.748 & 0.556 & 0.472 & 0.742 & 0.665\tabularnewline
\hline 
\end{tabular}
 \vspace{1mm}
\caption{Class existence  (higher is better)}\vspace{4mm}


\begin{tabular}{ccccccccccccccc}
\hline 
 & ceiling & floor & wall & window & chair & bed & sofa & table & tv & door & cabinet & furn & objs & mean\tabularnewline
\hline 
human & 0.374 & 0.762 & 0.989 & 0.739 & 0.633 & 0.3445 & 1.202 & 0.609 & 0.576 & 0.935 & 3.71 & 25.41 & 0.867 & 2.86\tabularnewline
rgbpn2pns & 0.433 & 0.877 & 0.915 & 1.107 & 0.894 & 0.656 & 0.93 & 0.99 & 0.754 & 1.385 & 0.872 & 1.194 & 1.16 & 0.936\tabularnewline
pns2pns & 0.365 & 0.782 & 0.711 & 0.89 & 0.854 & 0.67 & 0.835 & 0.954 & 0.593 & 1.028 & 1.07 & 1.814 & 0.882 & 0.881\tabularnewline
\hline 
\end{tabular}
 \vspace{1mm}
\caption{Class size (lower is better). \label{table:size} }\vspace{3mm}
\end{table*}

\begin{figure*}[t]
\vspace{-5mm}
\centering
    \includegraphics[width=0.97\linewidth]{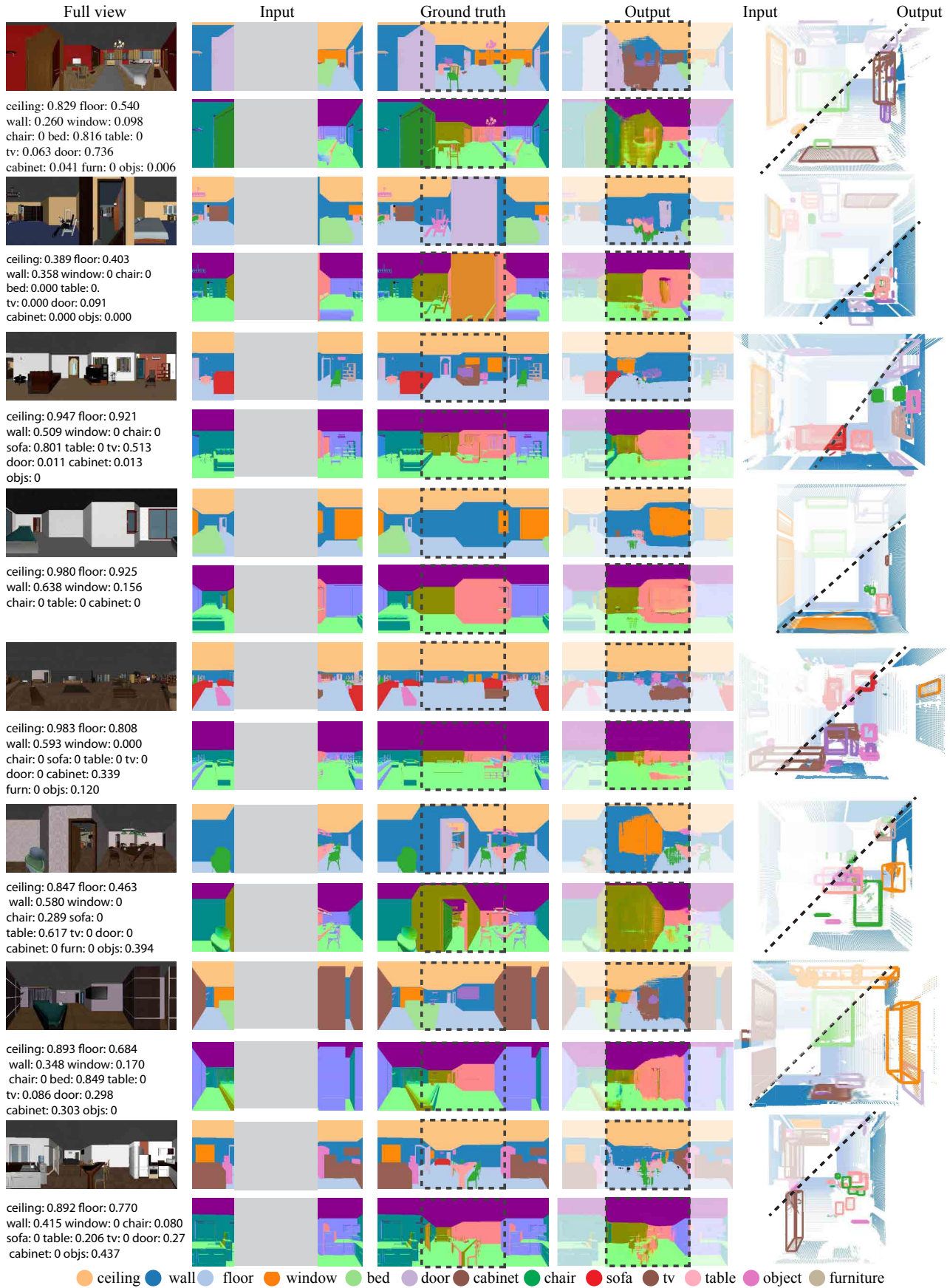}
    \caption{Additional result on SUNCG \cite{SSCNet} dataset.  Model: rgbpn2pns (s) }\label{fig:suncg_result}
\end{figure*}

\end{document}